%% file: main.tex
\documentclass{article}

\usepackage{microtype}
\usepackage{graphicx}
\usepackage{booktabs} 

\usepackage{hyperref}

\usepackage[accepted]{icml2023}

\usepackage{amsmath}
\usepackage{amssymb}
\usepackage{mathtools}
\usepackage{amsthm}
\usepackage{subcaption}
\usepackage{caption}
\usepackage{mathtools}
\usepackage{tabularx}
\usepackage{multirow}
\usepackage{multicol}
\usepackage{float}
\usepackage{listings}
\usepackage{algpseudocode}
\usepackage{xcolor}
\usepackage[capitalize,noabbrev]{cleveref}
\usepackage{pifont}
\newcommand{\xmark}{\ding{55}}%

\usepackage[symbol]{footmisc}

\theoremstyle{plain}

\theoremstyle{definition}

\theoremstyle{remark}

\usepackage[textsize=tiny]{todonotes}

\icmltitlerunning{DAG: Depth-Aware Guidance with Denoising Diffusion Probabilistic Models}

\begin{document}

\twocolumn[{
\icmltitle{DAG: Depth-Aware Guidance with Denoising Diffusion Probabilistic Models}



\icmlsetsymbol{equal}{*}
\icmlsetsymbol{corr}{$\dagger$}

\begin{icmlauthorlist}
    \icmlauthor{Gyeongnyeon Kim}{equal}
    \icmlauthor{Wooseok Jang}{equal}
    \icmlauthor{Gyuseong Lee}{equal}
    \icmlauthor{Susung Hong}{}
    \icmlauthor{Junyoung Seo}{}
    \icmlauthor{Seungryong Kim}{corr}\\
    Korea University, Seoul, Korea\\
    {\tt\small \{kkn9975,jws1997,jpl358,susung1999,se780,seungryong\_kim\}@korea.ac.kr}
\end{icmlauthorlist}



\icmlkeywords{Machine Learning, ICML}


\begin{figure}[H]
    \centering
    \hsize=\textwidth
    \captionsetup[subfigure]{labelformat=empty}
    \begin{subfigure}{\textwidth}
    \centering
    \lineskip=0pt
    \includegraphics[width=0.166\textwidth]{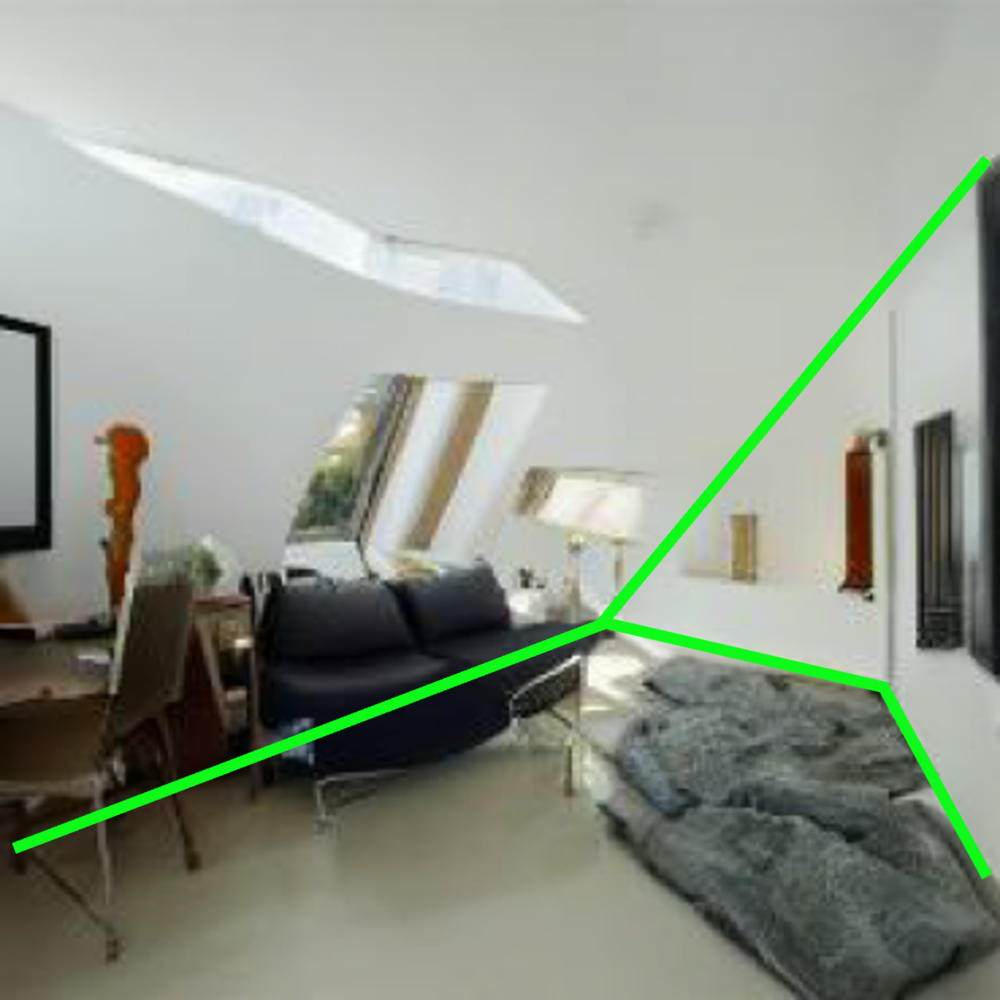}\hspace{-0.25em}
    \includegraphics[width=0.166\textwidth]{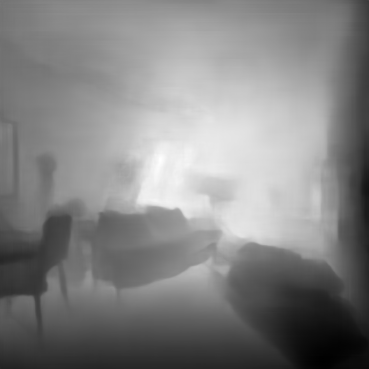}\hspace{-0.25em}
    \includegraphics[width=0.166\textwidth]{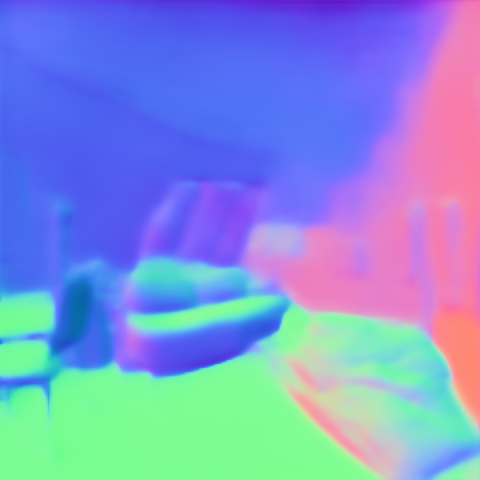}\hspace{-0.25em} 
    \includegraphics[width=0.166\textwidth]{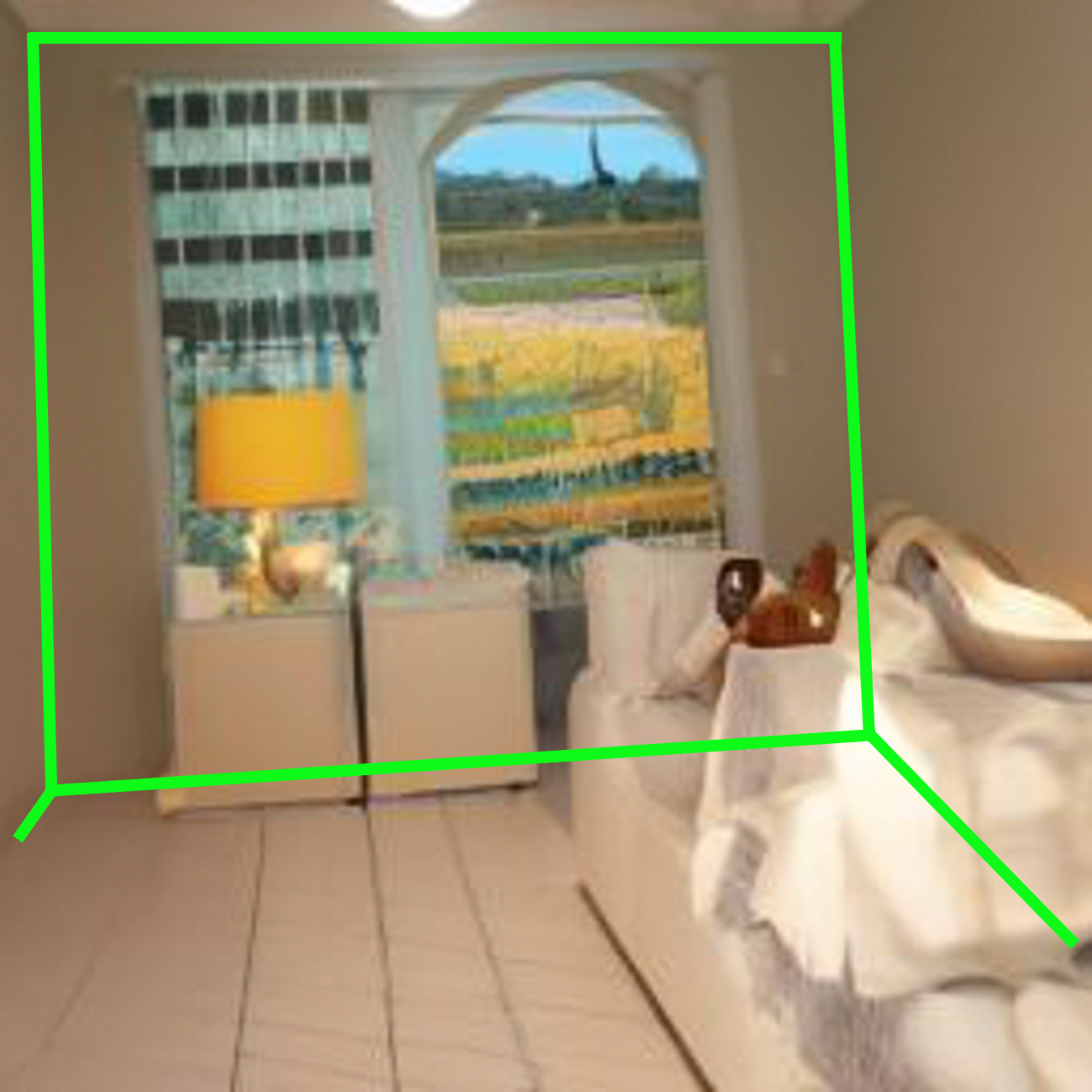}\hspace{-0.25em}
    \includegraphics[width=0.166\textwidth]{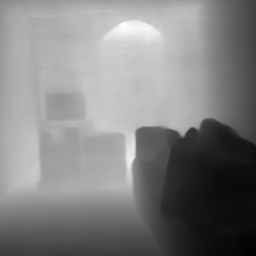}\hspace{-0.25em}
    \includegraphics[width=0.166\textwidth]{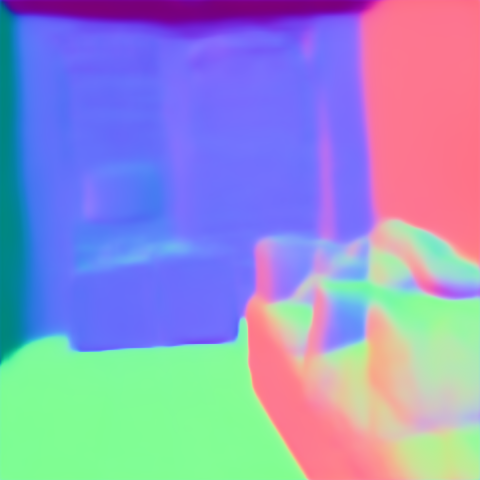} \\
    \includegraphics[width=0.166\textwidth]{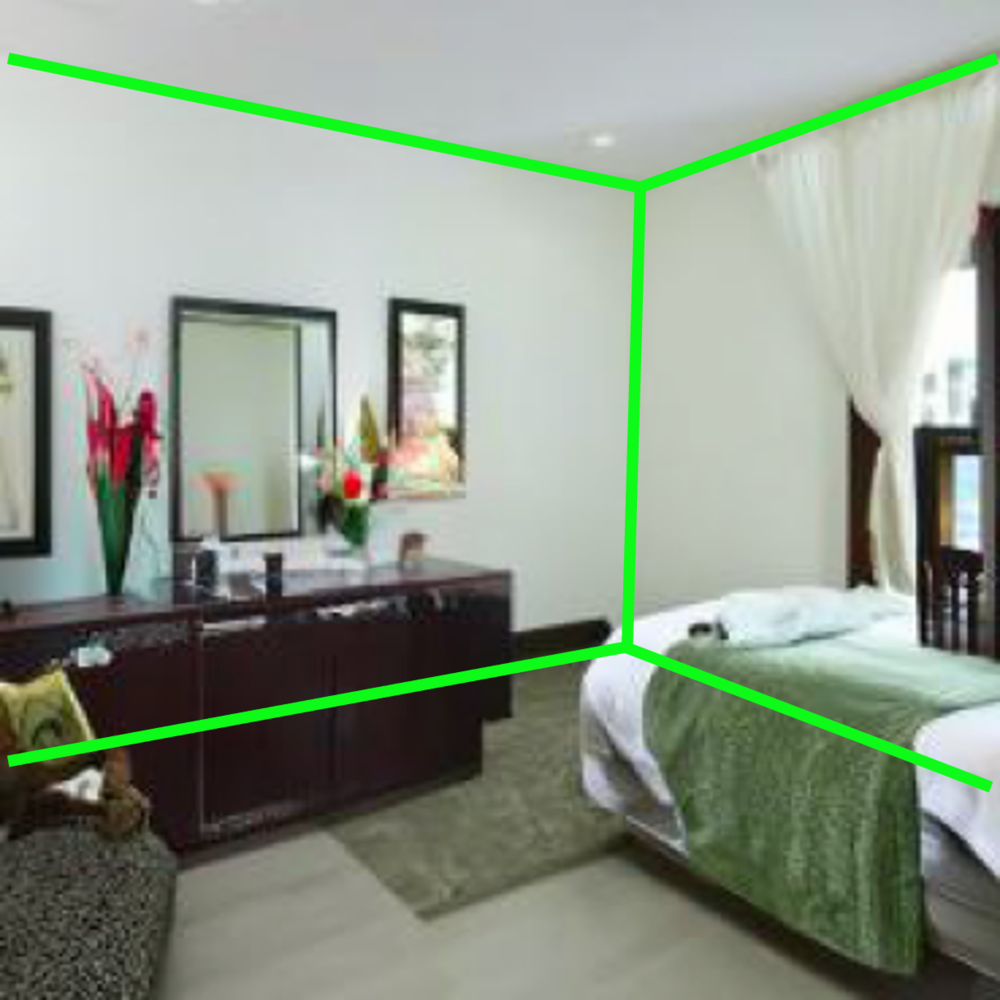}\hspace{-0.25em}
    \includegraphics[width=0.166\textwidth]{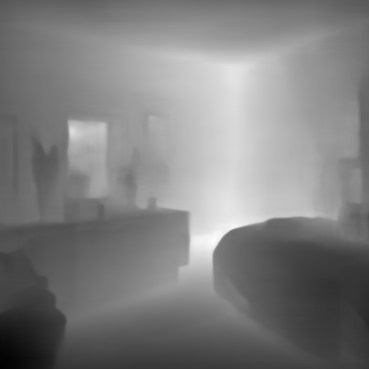}\hspace{-0.25em}
    \includegraphics[width=0.166\textwidth]{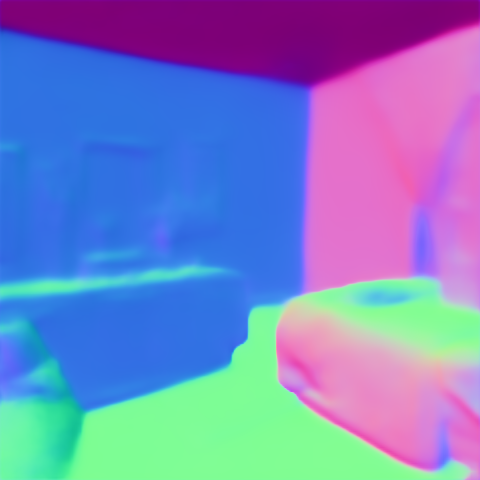}\hspace{-0.25em}
    \includegraphics[width=0.166\textwidth]{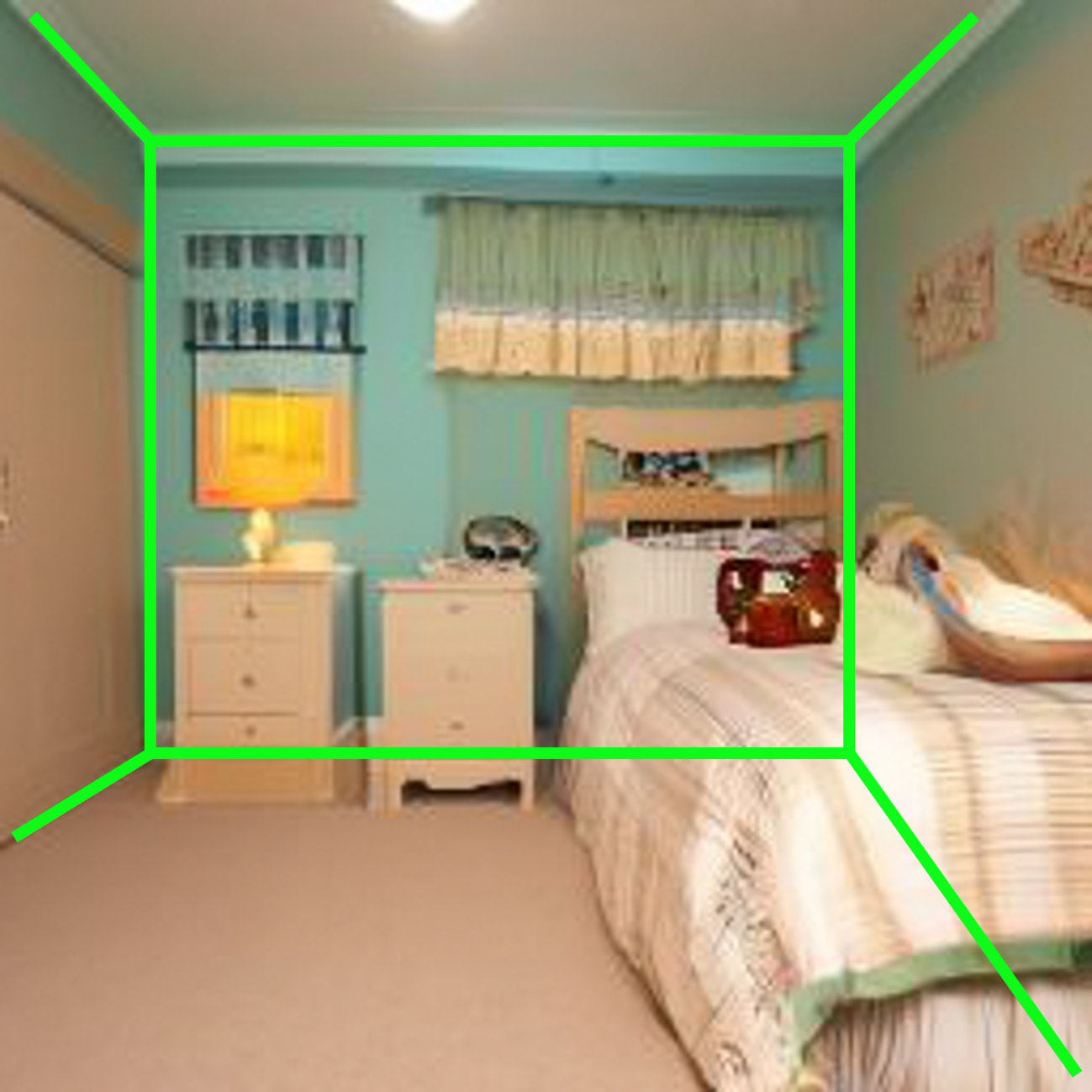}\hspace{-0.25em}
    \includegraphics[width=0.166\textwidth]{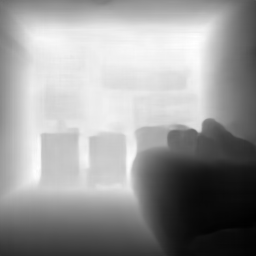}\hspace{-0.25em}
    \includegraphics[width=0.166\textwidth]{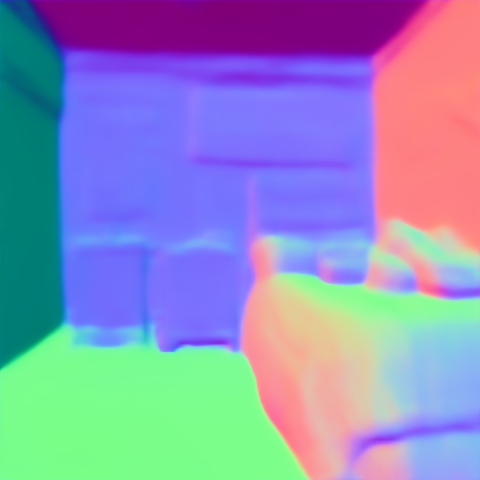}
    \end{subfigure}
    \vspace{-5pt}
    \caption{\textbf{Qualitative comparisons of synthesized images without guidance~\cite{dhariwal2021diffusion} (top) and with our depth-aware guidance (DAG), and their corresponding estimated depths~\cite{ranftl2021vision} and surface normals~\cite{Bae_2021_ICCV}.} Note that we highlight the scene layouts on the generated images. Our DAG helps better generate geometrically plausible images compared to the baseline~\cite{dhariwal2021diffusion}.}
    \label{fig:fig1}
\end{figure}

}]



\printAffiliationsAndNotice{\icmlEqualContribution} 

\input{latex/abstract.tex}

\input{latex/1_introduction.tex}
\input{latex/2_rel_work.tex}

\input{latex/3_method.tex}

\input{latex/4_experiment.tex}
\input{latex/5_conclusion.tex}

\bibliography{example_paper}
\bibliographystyle{icml2023}

\newpage
\appendix
\input{supple/supple.tex}

\end{document}

%% file: latex/abstract.tex
\begin{abstract}
    Generative models have recently undergone significant advancement due to the diffusion models. The success of these models can be often attributed to their use of guidance techniques, such as classifier or classifier-free guidance, which provide effective mechanisms to trade-off between fidelity and diversity. However, these methods are not capable of guiding a generated image to be aware of its geometric configuration, e.g., depth, which hinders their application to areas that require a certain level of depth awareness. To address this limitation, we propose a novel guidance method for diffusion models that uses estimated depth information derived from the rich intermediate representations of diffusion models. We first present label-efficient depth estimation framework using internal representations of diffusion models. Subsequently, we propose the incorporation of two guidance techniques based on pseudo-labeling and depth-domain diffusion prior during the sampling phase to self-condition the generated image using the estimated depth map. Experiments and comprehensive ablation studies demonstrate the effectiveness of our method in guiding the diffusion models towards the generation of geometrically plausible images.
\end{abstract}

%% file: latex/1_introduction.tex
\section{Introduction}

Diffusion models~\cite{sohl2015deep, ho2020denoising, nichol2021improved,rombach2022high} have recently received much attention and have shown remarkable generation quality. Such superior performance of diffusion models promoted many applications, such as text-guided image synthesis~\cite{nichol2021glide, ramesh2022hierarchical}, image restoration~\cite{saharia2022image, saharia2022palette, kawar2022denoising} and semantic segmentation~\cite{baranchuk2022labelefficient, asiedu2022decoder}.

Generally, generative models~\cite{goodfellow2020generative,ho2020denoising,farid2022perspective}, especially diffusion models, focus primarily on texture or appearance~\cite{tewari2022disentangled3d} but hardly consider shape geometry during the image generation process. As a result, as demonstrated in Fig.~\ref{fig:fig1}, conventional diffusion models often generate geometrically implausible images that contain ambiguous depth and cluttered object layouts. These synthesized samples with such unrealistic 3D geometry, whose deviation can be effectively captured by the depth map in 2D, are problematic in that they are not only visually unappealing but also unsuitable for downstream tasks, e.g., robotics and autonomous driving~\cite{wang2019pseudo,tateno2017cnn}.

While there have been some class-conditional guidance approaches~\cite{dhariwal2021diffusion,ho2021classifier} that drive the sampling process of diffusion models to a class-specific distribution, guiding the samples towards a geometrically plausible image distribution has received limited attention. In light of this, we propose a novel framework called Depth-Aware Guidance (DAG) that introduces depth awareness to diffusion models. Training diffusion models or depth predictors from scratch with depth-image pairs, as in conventional guidance~\cite{dhariwal2021diffusion,ho2021classifier}, can be challenging because it requires a lot of effort to annotate the ground-truth depth map and it takes tremendous time and computations to jointly train the models. To overcome these challenges, we employ a novel approach by utilizing the rich representations of a pretrained diffusion model to train depth predictors for the first time, thereby extending the knowledge of the representation capabilities of diffusion models in depth prediction tasks.

Furthermore, by leveraging label-efficient depth predictors, we propose a depth-aware guidance approach for image generation with the diffusion model. Specifically, we present two guidance strategies that effectively guide the geometric awareness: depth consistency guidance and depth prior guidance. The first strategy, depth consistency guidance, is motivated by consistency regularization~\cite{bachman2014learning,sajjadi2016regularization,laine2016temporal}. By treating the better prediction as a depth pseudo-label, depth consistency guidance guides the image toward improving the poor prediction. The second strategy, depth prior guidance, utilizes an additional pretrained diffusion U-Net as a prior network~\cite{graikos2022diffusion,poole2022dreamfusion} to provide guidance during the sampling process. This design explicitly injects depth information into the sampling process of diffusion models.

To evaluate our framework, we conduct experiments on indoor and outdoor scene datasets~\cite{yu2015lsun,Silberman:ECCV12} and propose new metric from the perspective of depth estimation tasks to capture geometric awareness. Our results, both qualitative and quantitative, demonstrate the superiority of our approach, and our extensive ablation studies validate its effectiveness. To the best of our knowledge, our work is the first attempt to utilize depth information during the sampling process to make image generation more aware of geometric configuration.

To sum up, we present the following key contributions:
\begin{itemize}
  \item We investigate the depth information contained in the learned U-Net representations of diffusion models.
  \item Based on the investigation, we propose a novel framework that can impose depth awareness of images generated from diffusion models. 
  \item The training of depth predictors in this framework can be done in a label-efficient manner by piggybacking on pretrained diffusion models.
  \item We propose novel guidance strategies that make use of consistency regularization and depth prior.     

\end{itemize}

%% file: latex/2_rel_work.tex
\section{Related Work}

\paragraph{Denoising diffusion probabilistic models.}
Diffusion models~\cite{sohl2015deep, ho2020denoising}, as the name states, model the reverse diffusion process, which gradually removes Gaussian noise from the latent variable. \cite{song2020score} have shown that diffusion models are equivalent to score-based generative models via stochastic differential equation interpretation. These models allow us to generate images from a complete Gaussian noise. Various attempts have been made to enhance the sample quality and sampling speed through different approaches~\cite{nichol2021improved,hong2022improving,dhariwal2021diffusion,rombach2022high,song2021denoising}. \cite{nichol2021improved} estimated the variance of reverse diffusion process and \cite{dhariwal2021diffusion} searched for an optimal U-shaped architecture for better sampling quality. \cite{song2021denoising} proposed non-Markovian diffusion process to reduce sampling steps whereas \cite{rombach2022high} leveraged diffusion process in the discrete latent space introduced in \cite{esser2020taming} for efficient computation and faster sampling. As a result of this advancement, diffusion models now outperform previous state-of-the-art models in various image generation tasks~\cite{nichol2021glide,seo2022midms,saharia2022photorealistic,ramesh2022hierarchical, rombach2022high,choi2021ilvr}. Additionally, text-to-image diffusion models~\cite{rombach2022high,saharia2022photorealistic,balaji2022ediffi} have been utilized as a strong image prior for text-to-3D generation~\cite{poole2022dreamfusion,lin2022magic3d} in recent studies.
\vspace{-10pt}

\paragraph{Sampling guidance for diffusion models.}
\cite{dhariwal2021diffusion} first proposed classifier guidance that leverages the gradient of an external classifier during the sampling process. By scaling the gradient, we can control the trade-off between the fidelity and diversity of the generated images. As an alternative without needing an external classifier, classifier-free guidance~\cite{ho2021classifier} has been proposed. During training, they randomly drop a certain percentage of class labels to obtain both conditional and unconditional models with shared parameters. With these two models, they get a similar effect on the classifier guidance. Applying these guidance leads to significant development in a conditional generation like text-to-image generation~\cite{nichol2021glide, rombach2022high, saharia2022photorealistic, ramesh2022hierarchical, gu2022vector} and semantic map-guided image generation~\cite{wang2022semantic, wang2022pretraining}. However, to our best knowledge, guidance for sampling with geometric awareness has been unexplored. 
\vspace{-10pt}

\paragraph{Internal representations in diffusion models.}
Recent studies~\cite{baranchuk2022labelefficient, asiedu2022decoder} have also analyzed and utilized the intermediate U-Net representations of diffusion models. \cite{baranchuk2022labelefficient} demonstrated that the U-Net representations of diffusion models at different layers contain semantic contexts of images and can predict semantic segmentation maps using only a shallow MLP and a scarcely-labeled dataset. Also, \cite{asiedu2022decoder} found similar phenomena that decoder pretraining of recovering the noised image like denoising autoencoders~\cite{vincent2008extracting} achieves high performance in label-efficient semantic segmentation task. For the first time, we propose to capture depth information using internal representations. 
\vspace{-10pt}

\paragraph{Monocular depth estimation.}
Monocular depth estimation, which aims to estimate a dense depth map from a single image~\cite{eigen2014depth, li2015depth,wang2015designing,kim2016,ummenhofer2017demon}, is an essential component of 3D perception. Since~\cite{eigen2014depth} proposed a deep learning-based approach for this task, several works~\cite{lee2019big,laina2016deeper,fu2018deep} have achieved significant progress due to the ability of CNNs to learn strong priors from images corresponding to the geometric layout. Following the recent success of Transformer~\cite{dosovitskiy2020image,vaswani2017attention} in vision tasks, \cite{ranftl2021vision, bhat2021adabins, li2022depthformer} employed a Transformer-based encoder to exploit a global receptive field. Despite notable improvements, recent works increase the model complexity and the reliance on large paired supervision.

%% file: latex/3_method.tex
\section{Preliminaries}
\paragraph{Denoising diffusion probabilistic models.}
\input{latex/FigureBox/figure2}

DDPM~\cite{ho2020denoising} is a kind of generative model that generates an image by iteratively denoising from Gaussian noise. In specific, given an input image $\mathbf{x}_0$ and an input noise $\epsilon\sim\mathcal{N}(\mathbf{0}, \mathbf{I})$, DDPM gradually adds noise according to a timestep $t$. For a pre-defined variance schedule $\beta_t$ for $t\in\{T, T-1, \ldots, 1\}$, we denote $\alpha_t$ and $\Bar{\alpha}_t$ as $1-\beta_t$ and $\Pi^t_{k=1} \alpha_k$, respectively. Then, given an arbitrary $t$, the forward process can be expressed in closed form~\cite{ho2020denoising}:
\begin{equation}
    \mathbf{x}_t = \sqrt{\Bar{\alpha}_t}\mathbf{x}_0 + \sqrt{1-\Bar{\alpha}_t}\mathbf{\epsilon}.
\end{equation}
As noted in Ho et al.~\cite{ho2020denoising}, we can re-weight the training objective of DDPM as:
\begin{equation}
    \mathcal{L}_\mathrm{simple}=\|\epsilon - \epsilon_\theta(\mathbf{x}_t)\|_2^2,
\end{equation}
where $\epsilon_\theta$ is a denoising U-Net~\cite{ho2020denoising} parameterized with $\theta$. Subsequently, the sampling process of DDPM is defined as the reverse of the forward process:
\begin{equation}
    \mathbf{x}_{t-1}\sim\mathcal{N}(\mu_{\theta}(\mathbf{x}_t), \sigma_t^2),
\label{eq:ddpm-sampling}
\end{equation}
where $\sigma_t^2$ is a constant only dependent on $t$, and $\mu_{\theta}$ is a prediction of the mean computed from the reparameterization of $\epsilon_\theta$. In some works~\cite{nichol2021improved, dhariwal2021diffusion}, the variance is also predicted and then Eq.~\ref{eq:ddpm-sampling} turns into:
\begin{equation}
    \mathbf{x}_{t-1}\sim\mathcal{N}(\mu_{\theta}(\mathbf{x}_t), \Sigma_{\theta}(\mathbf{x}_t)),
\label{eq:var-sampling}
\end{equation}
where $\Sigma_{\theta}$ is a prediction of variance in the reverse process from the model output. 
To formulate the task as conditional image generation with diffusion models, the need for guidance method is increasing. 

\vspace{-10pt}

\paragraph{Classifier guidance for diffusion model.}
By providing a way to trade off diversity and fidelity, classifier guidance~\cite{dhariwal2021diffusion} can improve the sample quality for image synthesis tasks. For a trained classifier $p_\psi(\mathbf{y}|\mathbf{x}_t)$, which receives the noised image $\mathbf{x}_t$ and predicts the class $\mathbf{y}$, a gradient with respect to $\mathbf{x}_t$ is calculated as:
\begin{equation}
    \nabla_{\mathbf{x}_{t}}\log(p_\psi(\mathbf{y}|\mathbf{x}_t)).
\end{equation}
Then, the sampling process with classifier guidance is defined as follows:
\begin{equation}
    \mathbf{x}_{t-1}\sim\mathcal{N}(\mu_{\theta}(\mathbf{x}_t)+w\Sigma\nabla_{\mathbf{x}_{t}}\log(p_\psi(\mathbf{y}|\mathbf{x}_t)), \Sigma_{\theta}(\mathbf{x}_t)).
\label{eq:cg}
\end{equation}
where $w>0$ is a scale of guidance. 

Even though this technique improves the sampling quality, no studies have attempted to extend it by leveraging dense predictors that incorporate geometric awareness.

\section{Methodology}
\subsection{Motivation and Overview}
Although the synthesized images by existing generative models such as GANs~\cite{goodfellow2014generative} or diffusion models~\cite{sohl2015deep, ho2020denoising} seem to be fairly plausible, they often lack geometric awareness. For example, the generated images by the existing models~\cite{farid2022perspective} often contain out-of-perspective or distortion of the layout, as depicted in Fig.~\ref{fig:qual_main}. This failure of considering geometric awareness can be an obstacle when applied to many downstream tasks~\cite{lin2013holistic,tateno2017cnn} that require geometrically realistic images. 

To overcome this limitation, we propose a novel framework, illustrated in Fig.~\ref{fig:main_figure}, to explicitly incorporate geometric information into the diffusion model as guidance. Specifically, to generate depth-aware images, we train depth predictors using internal representation~\cite{baranchuk2022labelefficient} with a small amount of depth-labeled data. Then, we guide the intermediate images using obtained depth map during the sampling process. 

In Sec.~\ref{sec:label_eff}, we demonstrate a way to acquire a depth map in a label-efficient way, and in Sec.~\ref{sec:guide}, we propose sampling methods to guide the generated images to have depth awareness. 

\subsection{Label-Efficient Training of Depth Predictors}
\label{sec:label_eff}

In order to generate depth-aware images with diffusion guidance in a straightforward way as in ~\cite{dhariwal2021diffusion}, we need either a large amount of image-depth pairs or an external large-scale depth estimation network trained on the noised images, both of which are challenging to acquire. To address this problem, we re-use the rich representations learned with DDPM that may contain depth information of images to estimate the depth.

\vspace{-10pt}
\paragraph{Network architecture.}
Recent research has shown that the internal features of the networks trained with diffusion models can encode semantic information~\cite{vincent2008extracting, baranchuk2022labelefficient, asiedu2022decoder}, and our contribution builds upon this by incorporating depth information into the framework.

To perform depth estimation in a label-efficient manner, we utilize a pixel-wise shallow Multi-Layer Perceptron (MLP) regressor that receives intermediate features from U-Net and estimates the depth values of the noisy input image. Specifically, we acquire the internal features $\mathbf{f}_t(k) \in \mathbb{R}^{C(k) \times H(k) \times W(k)}$ from the output of $k$-th decoder layer in the diffusion U-Net, where $C(k)$ denotes the channel dimension and $H(k)\times W(k)$ denotes the spatial resolution of the $k$-th layer of the U-Net decoder. 
Subsequently, we form the depth map by querying the MLP blocks pixel-by-pixel, where the depth map can be formulated as:
\begin{equation}
   \mathbf{d}_{t}(k) = \text{MLP}(\mathbf{f}_{t}(k)). \label{eq:mlp}
\end{equation}

Also, similar to \cite{baranchuk2022labelefficient}, it is generally better to use more features from different U-Net layers than only using the feature from one layer. Therefore we extract more features from many layers and concatenate them in a channel dimension, which can be formulated as:
\begin{equation}
    \mathbf{g}_t=[\mathbf{f}_t(1);\mathbf{f}_t(2);\cdots;\mathbf{f}_t(d)],
    \label{eq:concat}
\end{equation}
where $[\cdot\, ; \cdot]$ is a channel-wise concatenation operation between features, and $d$ is the total number of selected layers. Then, we pass them to the pixel-wise MLP depth predictor, and the generalization of Eq.~\ref{eq:mlp} in terms of Eq.~\ref{eq:concat} can be stated as:
\begin{equation}
    \mathbf{d}_{t} = \text{MLP}(\mathbf{g}_t).
\end{equation}

We modify the pixel-wise depth predictor by appending an additional time-embedding block to the input in order to predict the depth map at arbitrary timesteps. Therefore, we can use this predictor throughout the entire sampling process. We follow the time-embedding module of diffusion U-Net. Applying it to Eq.~\ref{eq:mlp}, we can rewrite the equation as:
\begin{equation}
    \mathbf{d}_{t} = \text{MLP}(\mathbf{g}_t, t).
\end{equation}

\input{latex/FigureBox/seq_simple.tex}

\vspace{-10pt}
\paragraph{Loss function.}
We train the depth estimator only with the frozen features from the diffusion U-Net by using the ground-truth depth map $\mathbf{y}$ with L1 loss as
\begin{equation}
   \mathcal{L}_\mathrm{depth}=\|\mathbf{d}_{t} - \mathbf{y}\|_1.
   \label{eq:depthloss}
\end{equation}

This whole procedure allows us to achieve reasonable label-efficient prediction performance in the depth domain, as demonstrated in Fig.~\ref{fig:abl_depth}(b). The depth prediction scheme allows us to predict the depth map not only for any input images but also for the intermediate images under generation in arbitrary sampling steps, as shown in Fig.~\ref{fig:seq}, since the representations of diffusion models are inherently learned with the timesteps.

\subsection{Depth Guided Sampling for Diffusion Model} \label{sec:guide}
To ensure that the generated images yield plausible depth maps, we encourage the predicted depth maps to be accurate during the sampling process. To this end, we propose two different guidance techniques that use the representations of the denoising U-Net. Utilizing these techniques during the sampling process regularizes the model to generate images that are aware of the dense map prior. Specifically, we use the aforementioned efficiently-trained depth predictors in Sec.~\ref{sec:label_eff}, which can predict the depth maps from images at any sampling steps.

Unfortunately, because of the absence of a pre-determined label, we cannot compute loss using this label and cannot give guidance during the sampling process. Therefore we build two alternative loss functions that can act as guidance constraints, which we discuss in detail in the following sections. Considering Eq.~\ref{eq:cg}, The general form of guidance equation is formulated by:
\begin{equation}
    \mathbf{x}_{t-1} \sim \mathcal{N}(
        \mu_\theta(\mathbf{x}_t) -
        \omega\nabla_{\mathbf{x}_t}{\mathcal{L}_\mathrm{depth}}, 
        \Sigma_\theta(\mathbf{x}_t)
    ).
    \label{eq:cg-depth}
\end{equation}

\vspace{-10pt}

\input{latex/FigureBox/abl_depth_performance.tex}

\paragraph{Depth consistency guidance (DCG).}
One naive approach for providing depth guidance to the images in the absence of ground truth depth information is to use pseudo-labeling~\cite{sohn2020fixmatch, lee2013pseudo}. However, it is challenging to generate a confident prediction that can be pseudo-label. Our first guidance method is inspired by FixMatch~\cite{sohn2020fixmatch}, which combines pseudo-labeling~\cite{lee2013pseudo} and consistency regularization~\cite{bachman2014learning,sajjadi2016regularization}. It boosts performance by considering the weakly-augmented labels as pseudo-label. Our hypothesis derives from this design: we argue that the richer the representation is, the more faithful produced depth map becomes. Therefore, we regard the predictions from the aggregation of more feature blocks to be more informative, which makes them suitable to be treated as robust predictions (Fig.~\ref{fig:seq}), i.e., pseudo-labels. We refer to these predictions that use multiple feature blocks as "strong branch predictions", and those that utilize relatively fewer features as "weak branch predictions".

\input{latex/FigureBox/figure3.tex}

In specific, we use the features $\mathbf{g}^{W}=[\mathbf{f}_t(6)]$ for weak branch features, and $\mathbf{g}^{S}=[\mathbf{f}_t(2); \mathbf{f}_t(4); \mathbf{f}_t(5); \mathbf{f}_t(6); \mathbf{f}_t(7)]$ for strong branch features.
To account for the different channel dimensions of these two aggregated features, we design two asymmetric predictors: $\text{MLP-S}$ and $\text{MLP-W}$. The first predictor receives more features from the U-Net block, while the second one receives fewer features, and we train them together. We feed the collected features to these MLPs and obtain the depth map predictions:
\begin{equation}
    \mathbf{d}^{S}_t = \text{MLP-S}(\mathbf{g}^{S}_t, t), \quad \mathbf{d}^{W}_t = \text{MLP-W}(\mathbf{g}^{W}_t, t).
\end{equation}
As stated above, we treat $\mathbf{d}^{S}_t$ as a pseudo-label and $\mathbf{d}^{W}_t$ as a prediction then compute the loss using a consistency loss between two predicted dense maps. We apply the stop-gradient operation to the strong features, preventing the strong prediction from learning the weak prediction~\cite{chen2021exploring, sohn2020fixmatch}. The gradient of the loss with respect to $\mathbf{x}$ flows through the diffusion U-Net and guides the sampling process
as done in~\cite{dhariwal2021diffusion}. This process can be formulated as
\begin{equation}
    \mathcal{L}_\mathrm{dc} = \|\mathrm{stopgrad}(\mathbf{d}^{S}_t)-\mathbf{d}^{W}_t\|^2_2,
\end{equation}
where $\mathrm{stopgrad}$ denotes the stop-gradient operation.

Finally, we can guide the generation process by using this gradient with Eq.~\ref{eq:cg-depth}, and it can be formulated as:
\begin{equation}
\mathbf{x}_{t-1}
    \sim
    \mathcal{N}(
        \mu_\theta(\mathbf{x}_t) -
        \omega_\mathrm{dc}\nabla_{\mathbf{x}_t}{\mathcal{L}_\mathrm{dc}}, 
        \Sigma_\theta(\mathbf{x}_t)),
        \label{eq:dcg}  
\end{equation}
where $\omega_\mathrm{dc}$ denotes the guidance scale. 
\vspace{-10pt}

\input{latex/FigureBox/pointcloud.tex}

\paragraph{Depth prior guidance (DPG).}
We also propose another guidance method, which we call depth prior guidance, to inject depth prior into the sampling process. The pretrained diffusion model can effectively refine the noised distributions to realistic distributions~\cite{meng2021sdedit}, or it can help to optimize the noised initialization of the data to match with the real data by utilizing the knowledge of the diffusion model~\cite{graikos2022diffusion,poole2022dreamfusion}.
Therefore we train another small-resolution diffusion U-Net $\epsilon_\phi$ on the depth domain and use it as our prior for the second guidance method. As described in Sec.~\ref{sec:label_eff}, we can extract the features from the decoder part of the image-generating U-Net to estimate the corresponding depth map using MLP depth predictor. Utilizing the prior diffusion model, we inject noise to the depth prediction $\mathbf{d}^{S}_0$ using a forward process of diffusion like:
\begin{equation}
    \mathbf{d}^{S}_\tau = \sqrt{\Bar{\alpha}_\tau}\mathbf{d}^{S}_0 + \sqrt{1-\Bar{\alpha}_\tau}\eta, \quad \eta\sim\mathcal{N}(\mathbf{0}, \mathbf{I}),
\end{equation}
where $\tau$ is the timestep that is used in a prior diffusion model.
After adding noise to the depth prediction, we feed it to our prior network to estimate the added noise. Then we calculate the gradient of the mean-squared error between the added noise and the predicted noise concerning $x$. This process is then defined as:
\begin{equation}
   \mathcal{L}_\mathrm{dp}=\|\eta - \epsilon_\phi(\mathbf{d}^{S}_\tau)) \|^2_2.
\end{equation}
Treating the gradient of the above loss as an external classifier gradient like~\cite{dhariwal2021diffusion}, we can guide the generated image to match with the depth prior like:
\begin{equation}
\mathbf{x}_{t-1}
    \sim
    \mathcal{N}(
        \mu_\theta(\mathbf{x}_t) -
    \omega_\mathrm{dp}\nabla_{\mathbf{x}_t}\mathcal{L}_\mathrm{dp}, 
        \Sigma_\theta(\mathbf{x}_t)),
        \label{eq:dpg}
\end{equation}
where $\omega_\mathrm{dp}$ denotes the guidance scale.
\vspace{-10pt}

\paragraph{Overall guidance.}
By integrating the proposed DCG and DPG, we can guide the sampled image using both. Using Eq.~\ref{eq:dcg} and \ref{eq:dpg}, our overall sampling can be written as:
\begin{equation}
\begin{aligned}
\mathbf{x}_{t-1}
    \sim
    \mathcal{N}(\mu_\theta(\mathbf{x}_t)
    - \omega_\mathrm{dc}\nabla_{\mathbf{x}_t}{\mathcal{L}_\mathrm{dc}}-\omega_\mathrm{dp}\nabla_{\mathbf{x}_t}\mathcal{L}_\mathrm{dp}, \Sigma_\theta(\mathbf{x}_t)).
    \label{eq:overall}
\end{aligned}
\end{equation}

%% file: latex/FigureBox/figure2.tex
\begin{figure*}[t]
\begin{center}
\includegraphics[width=1.0\linewidth]{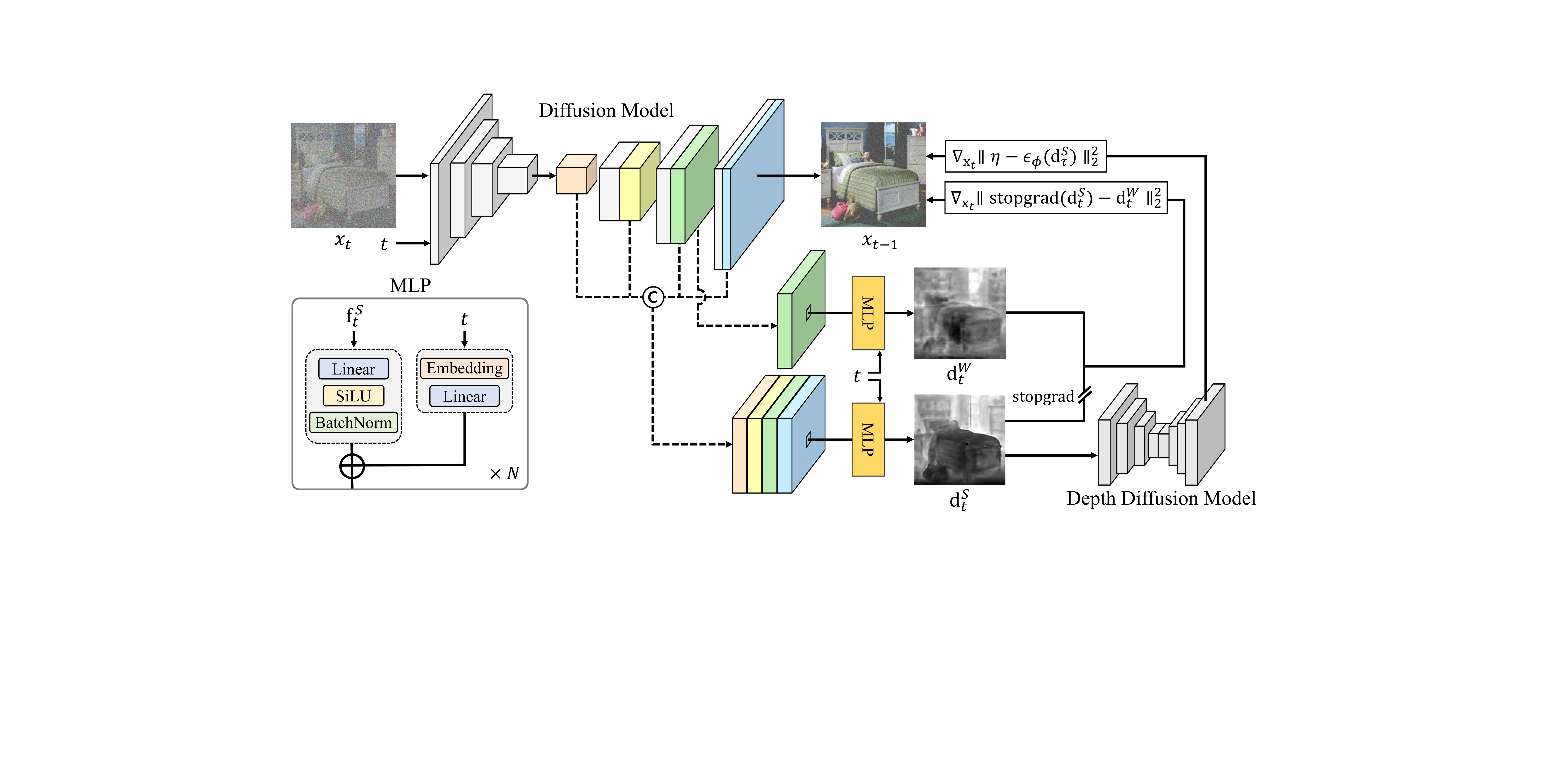}
\end{center}
\caption{\textbf{Overview of our framework.} First of all, we train asymmetric pixel-wise depth predictor conditioned on a timestep with respect to pretrained diffusion models in a label-efficient manner. Then, we apply two guidance strategies. First, we extract strong and weak depth maps with this predictor from DDPM network and give a depth consistency guidance. Next, giving the depth map from the strong branch as an input, the pretrained depth diffusion model is utilized to push the model prior into intermediate images to be depth-aware.}
\label{fig:main_figure}
\vspace{-10pt}
\end{figure*}

%% file: latex/FigureBox/seq_simple.tex
\begin{figure}[!t]

    \centering
    \small
    \setlength\tabcolsep{0.8pt}
    {
    \renewcommand{\arraystretch}{0.5}
    \resizebox{\columnwidth}{!}{%
    \begin{tabular}{ccccc}
         $t$=800 & $t$=600 & $t$=400 & $t$=200 & $t$=0 \\
      \includegraphics[width=0.165\linewidth]{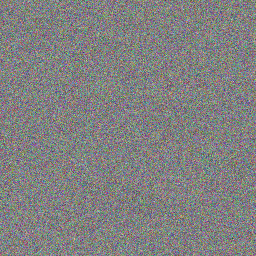}&
      \includegraphics[width=0.165\linewidth]{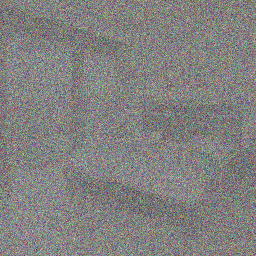}&
      \includegraphics[width=0.165\linewidth]{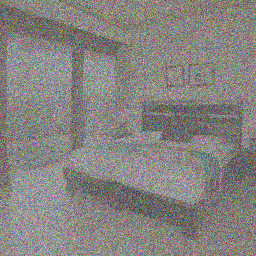}&
      \includegraphics[width=0.165\linewidth]{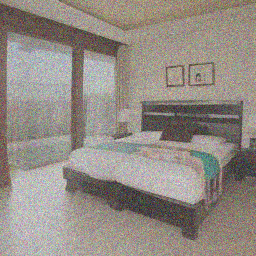}&
      \includegraphics[width=0.165\linewidth]{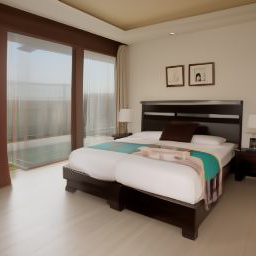} \\
      \includegraphics[width=0.165\linewidth]{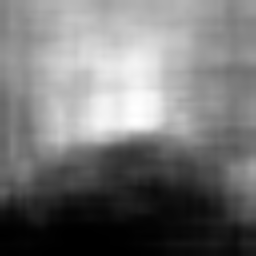}&
      \includegraphics[width=0.165\linewidth]{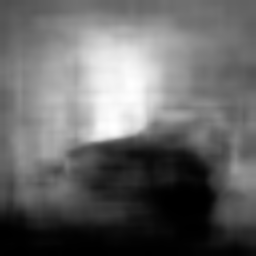}&
      \includegraphics[width=0.165\linewidth]{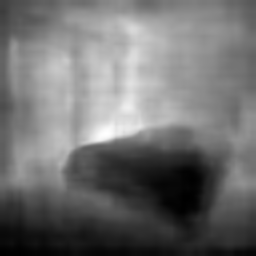}&
      \includegraphics[width=0.165\linewidth]{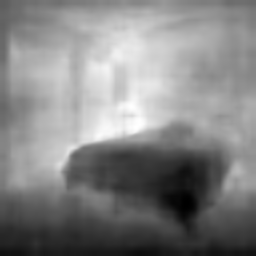}&
      \includegraphics[width=0.165\linewidth]{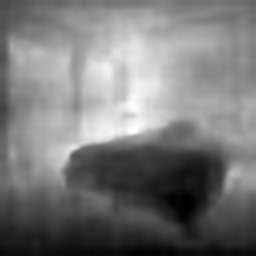}\\ 
      \includegraphics[width=0.165\linewidth]{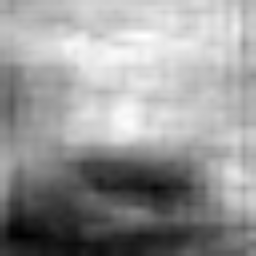}&    \includegraphics[width=0.165\linewidth]{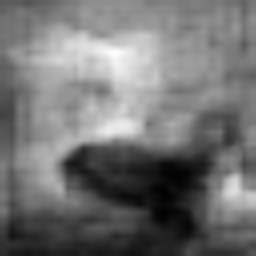}&
      \includegraphics[width=0.165\linewidth]{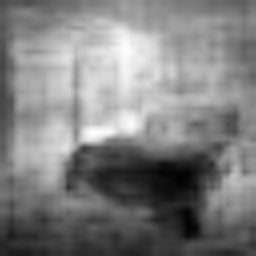}&
      \includegraphics[width=0.165\linewidth]{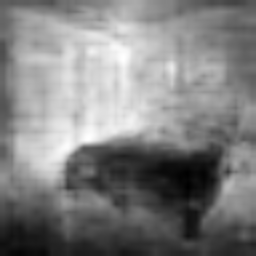}&
      \includegraphics[width=0.165\linewidth]{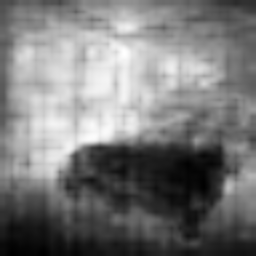}\\
    \end{tabular} }
    }
    \vspace{-5pt}
    \caption{\textbf{Visualizations of the sampling process of our framework: } (from top to bottom) predicted images, depth predictions from the strong-branch predictor ($\mathbf{d}^{S}_t$), and depth predictions from the weak-branch predictor ($\mathbf{d}^{W}_t$). As exemplified, the strong-branch predictor gives robust depth predictions even at the early stage. }
    \label{fig:seq} \vspace{-10pt}
    \end{figure}

%% file: latex/FigureBox/abl_depth_performance.tex
\begin{figure}[t]

\centering
  \begin{subfigure}[t]{0.49\linewidth}
    \centering
    \includegraphics[width=1\linewidth]{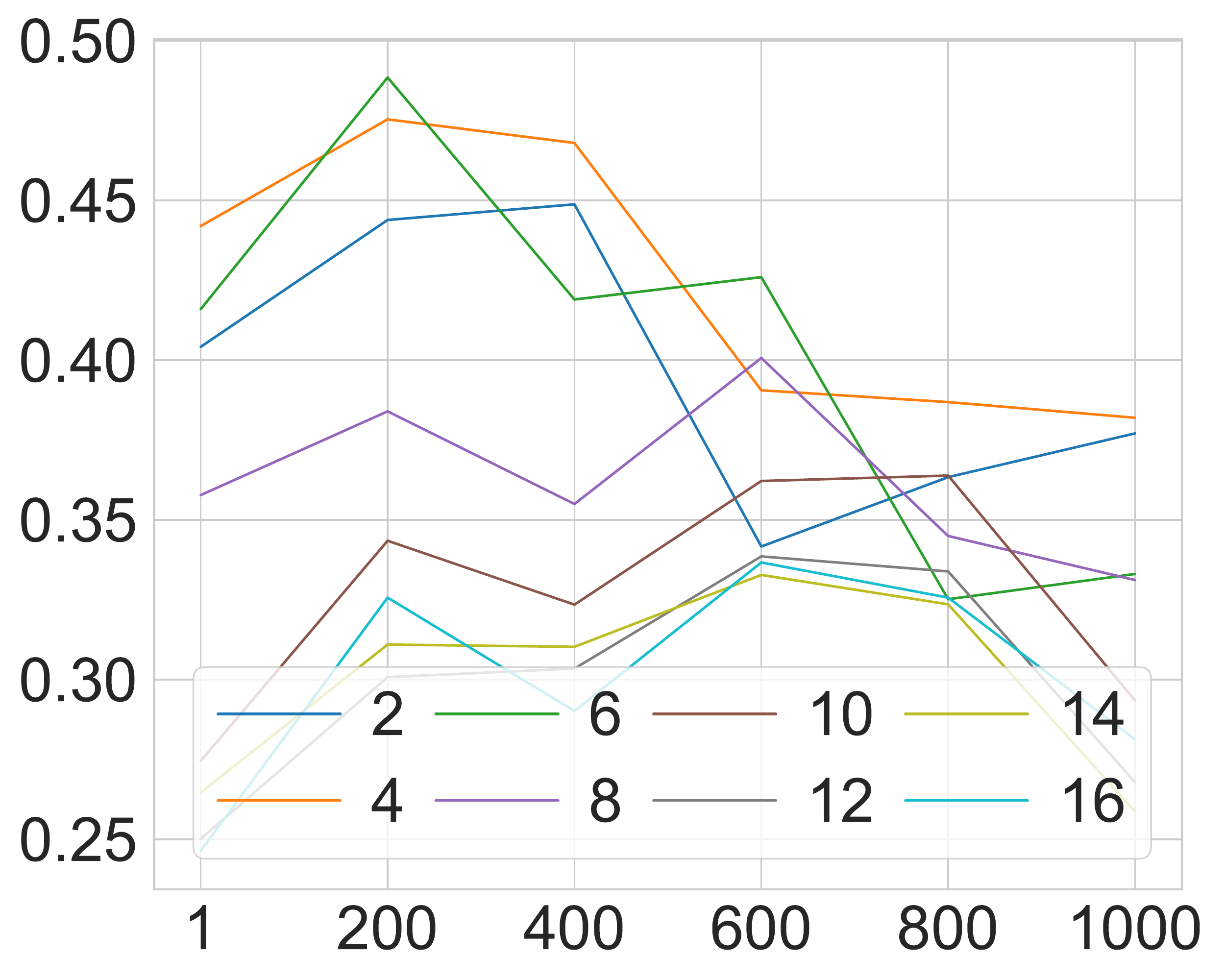}
    \caption{}

  \end{subfigure}
  \begin{subfigure}[t]{0.49\linewidth}
    \centering 
    \includegraphics[width=1.0\linewidth]{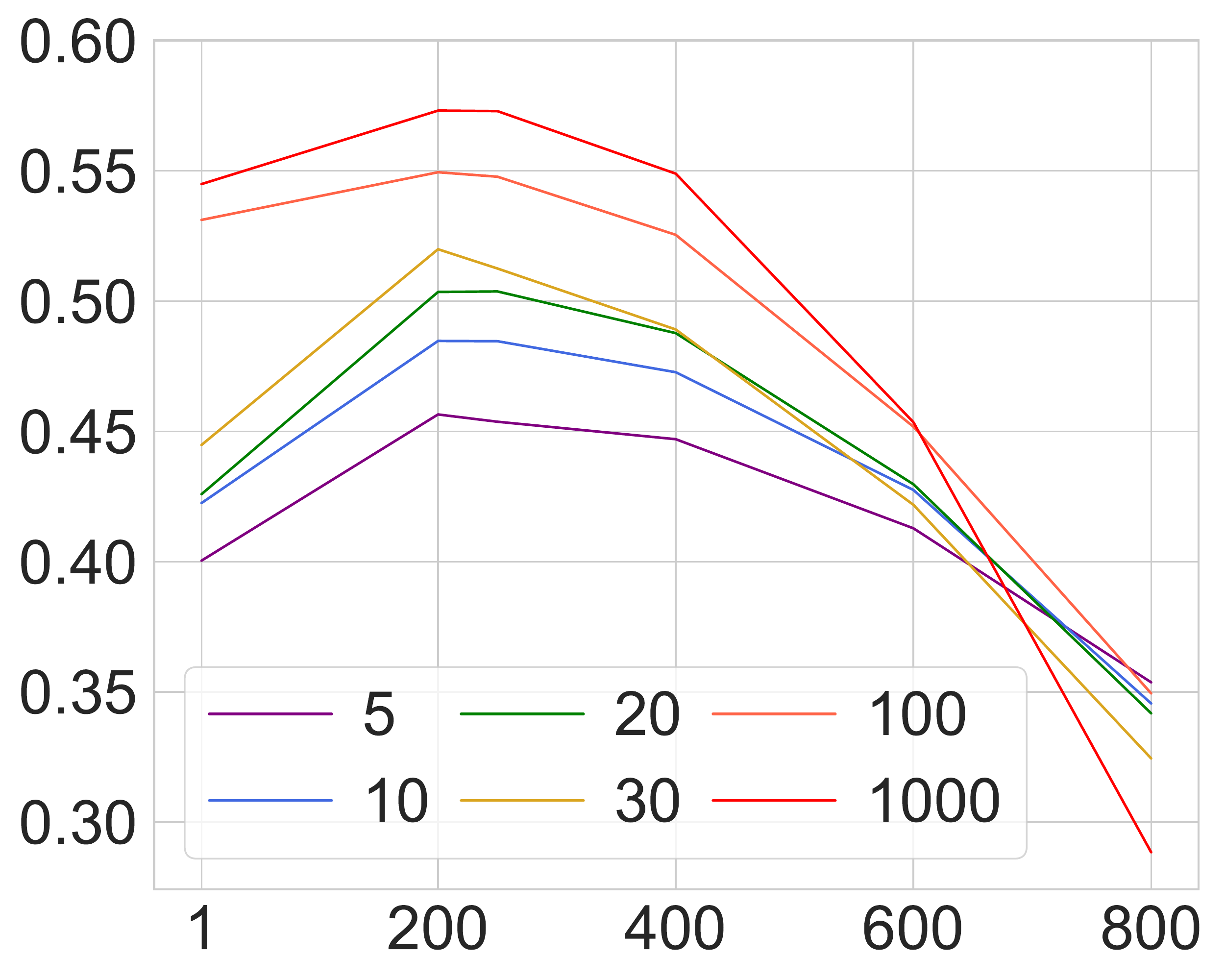}
    \caption{}

\end{subfigure}
\vspace{-5pt}
\caption{\textbf{Quantitative comparisons of depth prediction performance.} Evaluation of depth estimation performance for varying timesteps according to different U-Net blocks and the number of training images. The x-axis denotes timesteps and the y-axis denotes the accuracy of depth estimation. (a) Comparison of depth prediction accuracy according to different U-Net blocks. Each line denotes the layer from which the feature map is extracted in the decoder of U-Net. (b) Comparison of depth prediction accuracy with the number of images shown in training. For (b), each line denotes the number of samples shown.}
\vspace{-10pt}
\label{fig:abl_depth}
\end{figure}

%% file: latex/FigureBox/figure3.tex
\begin{figure*}[t]
    \centering
    \setlength\tabcolsep{0.8pt}
    {
    \renewcommand{\arraystretch}{0.5}
    \resizebox{\linewidth}{!}{%
    \begin{tabular}{cccccccc}
      \includegraphics[width=0.12\linewidth]{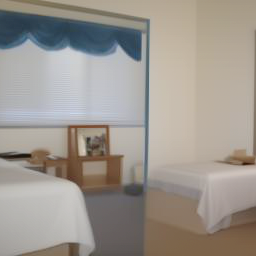}&
      \includegraphics[width=0.12\linewidth]{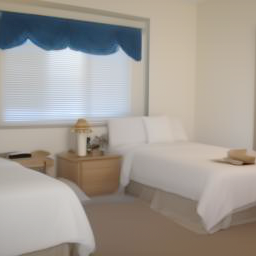}&
      \includegraphics[width=0.12\linewidth]{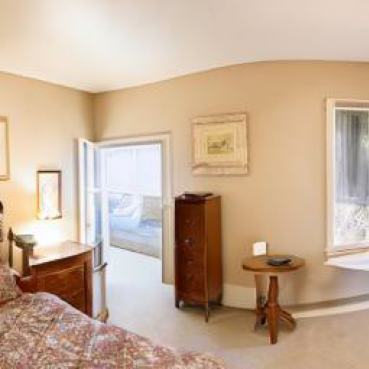}&
      \includegraphics[width=0.12\linewidth]{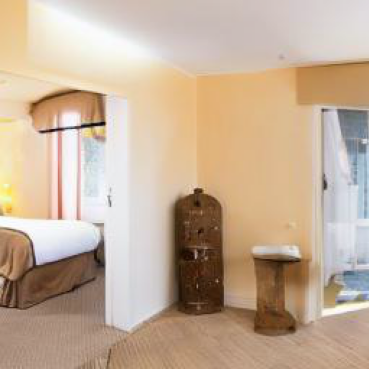}&
      \includegraphics[width=0.12\linewidth]{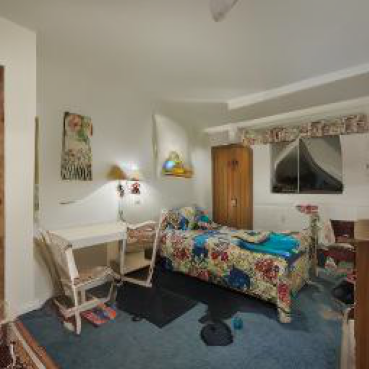}&
      \includegraphics[width=0.12\linewidth]{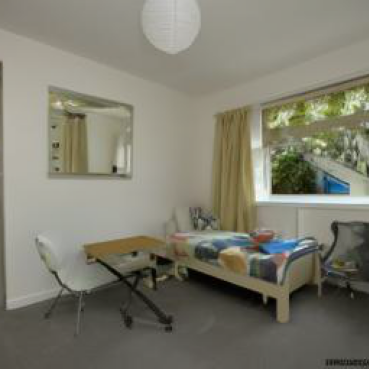}&
      \includegraphics[width=0.12\linewidth]{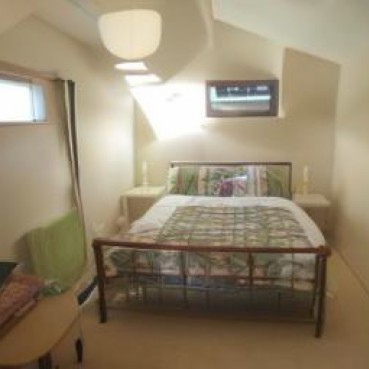}&
      \includegraphics[width=0.12\linewidth]{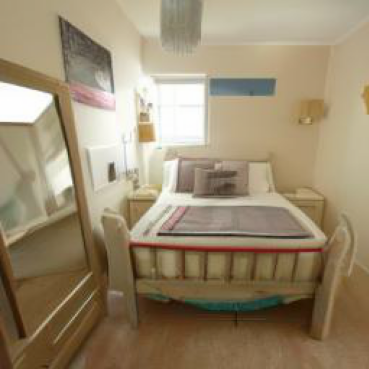}\\

      \includegraphics[width=0.12\linewidth]{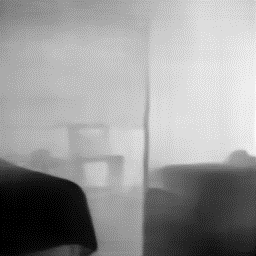}&
      \includegraphics[width=0.12\linewidth]{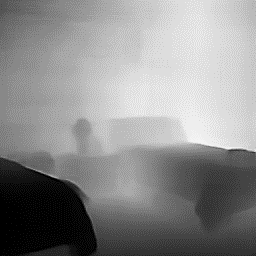}&
      \includegraphics[width=0.12\linewidth]{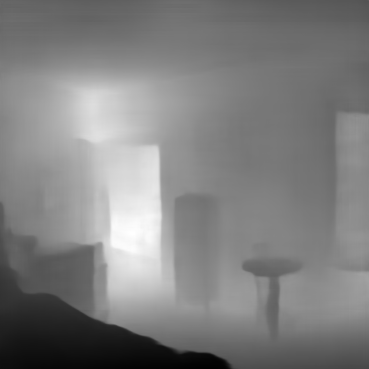}&
      \includegraphics[width=0.12\linewidth]{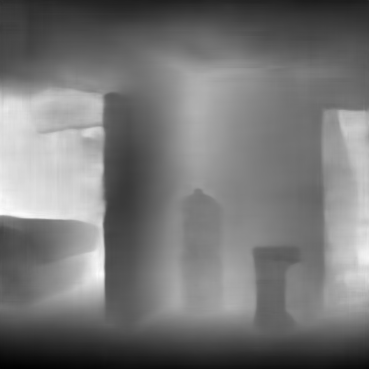}&
      \includegraphics[width=0.12\linewidth]{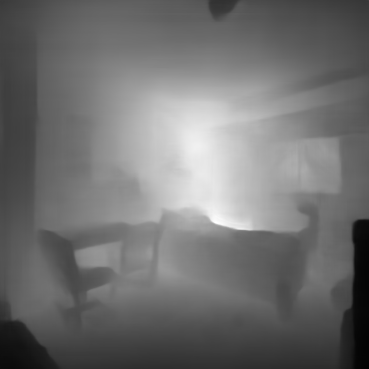}&
      \includegraphics[width=0.12\linewidth]{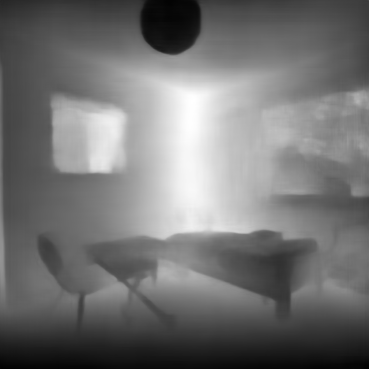}&
      \includegraphics[width=0.12\linewidth]{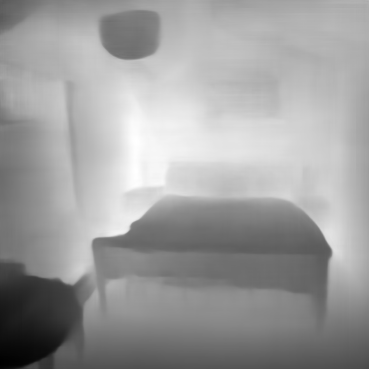}&
      \includegraphics[width=0.12\linewidth]{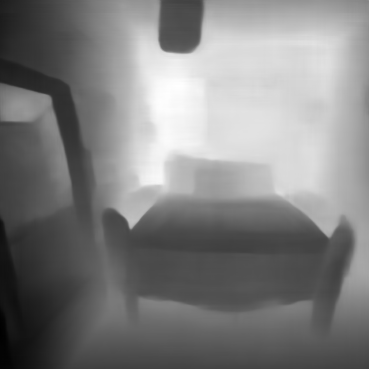}\\

     \includegraphics[width=0.12\linewidth]{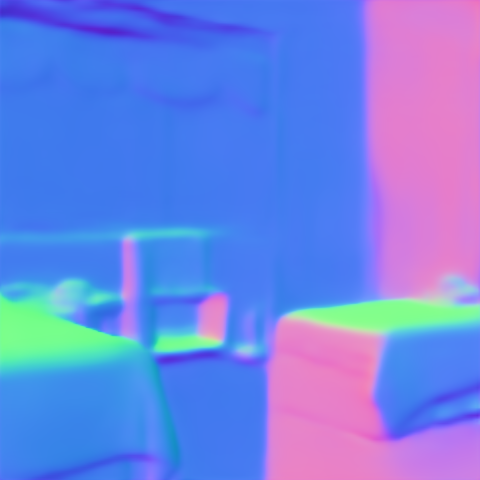}&
      \includegraphics[width=0.12\linewidth]{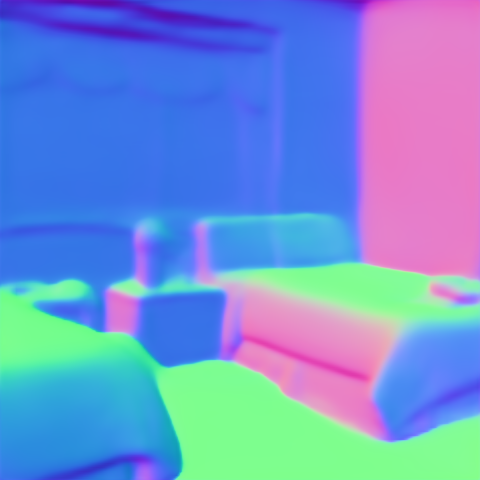}&
      \includegraphics[width=0.12\linewidth]{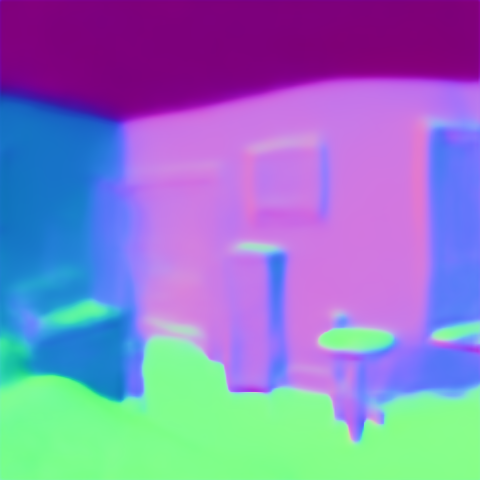}&
      \includegraphics[width=0.12\linewidth]{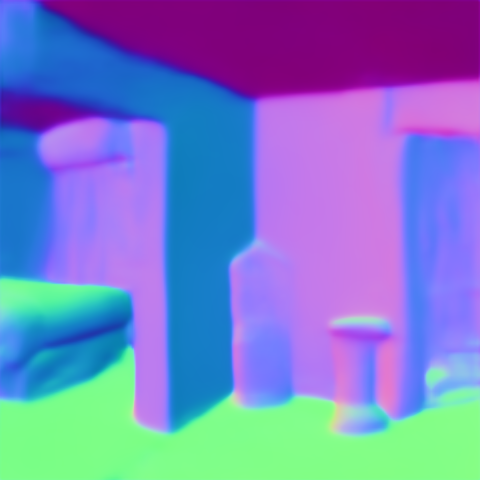}&
      \includegraphics[width=0.12\linewidth]{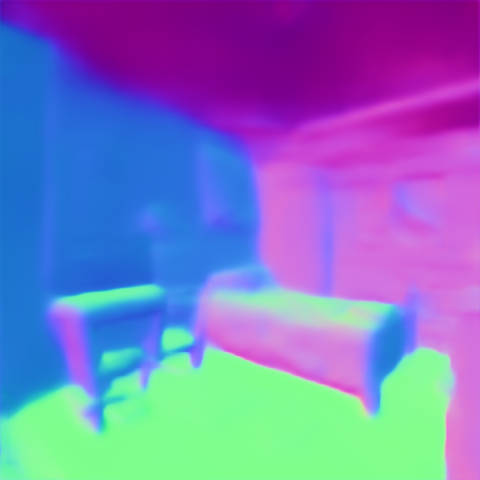}&
      \includegraphics[width=0.12\linewidth]{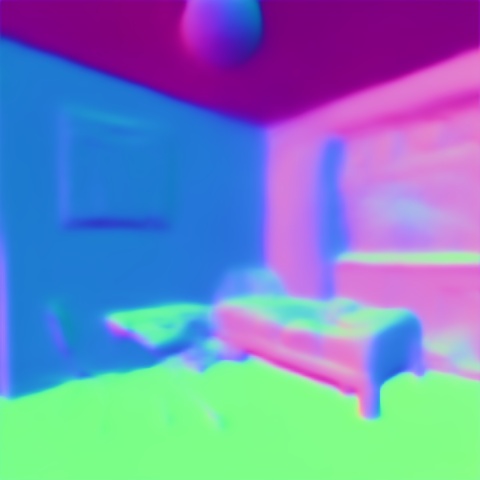}&
      \includegraphics[width=0.12\linewidth]{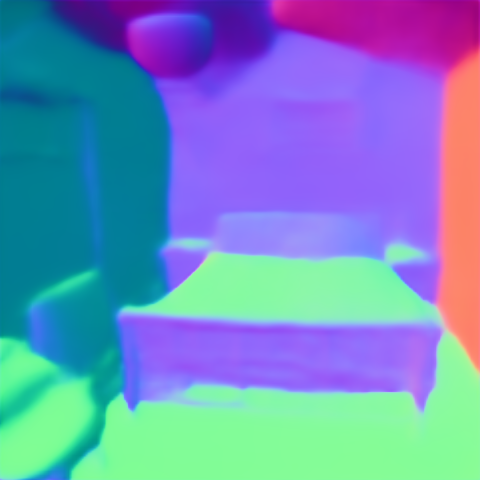}&
      \includegraphics[width=0.12\linewidth]{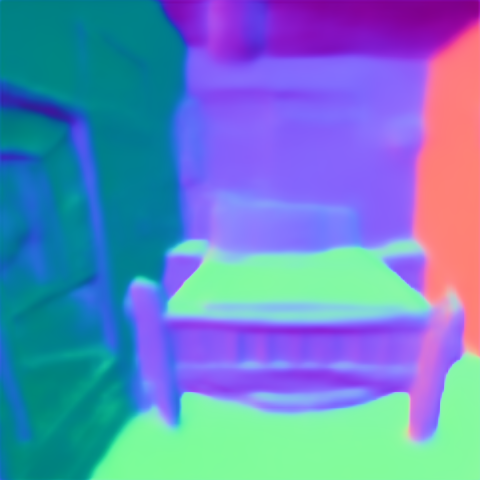}\\
      (a) & (b) & (c) & (d) & (e) & (f) & (g) & (h)\\
    \end{tabular} }
    }
    \vspace{-5pt}
    \caption{\textbf{Qualitative comparison on LSUN bedroom datasets~\cite{yu2015lsun}.}  We visualize the generated images (top) without guidance ((a), (c), (e), (g)) and with depth-aware guidance ((b), (d), (f), (h)), and their corresponding depths~\cite{ranftl2021vision} (middle) and surface normals~\cite{Bae_2021_ICCV} (bottom). }
    \label{fig:qual_main}
    \vspace{-10pt}
    \end{figure*}

%% file: latex/FigureBox/pointcloud.tex
\begin{figure*}[t]
\centering
\hsize=\linewidth
\captionsetup[subfigure]{labelformat=empty}
\begin{subfigure}{.49\textwidth}
\centering
\lineskip=0pt
      \includegraphics[width=0.24\textwidth]{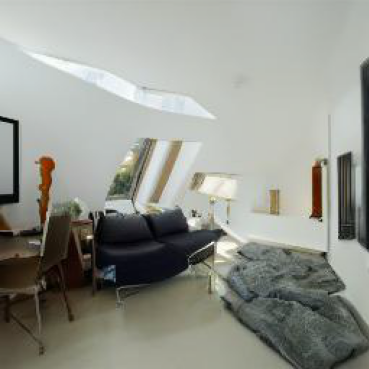}
      \includegraphics[width=0.24\textwidth]{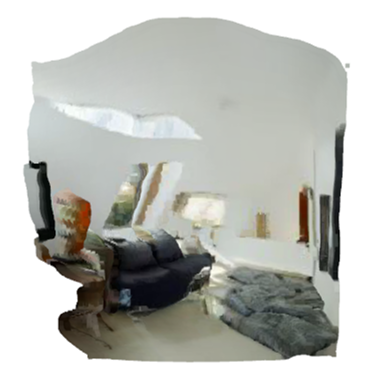}
      \includegraphics[width=0.24\textwidth]{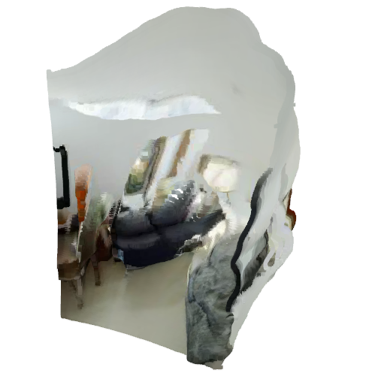}
      \includegraphics[width=0.24\textwidth]{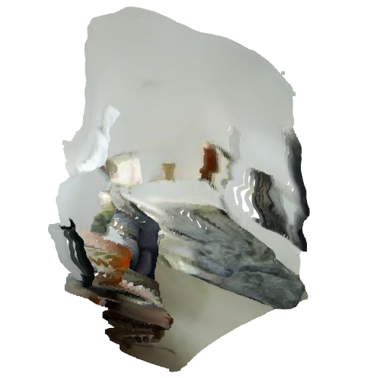}\\
      
      \includegraphics[width=0.24\textwidth]{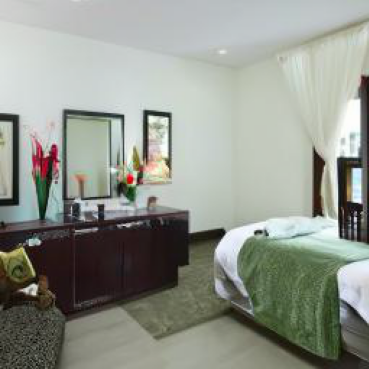}
      \includegraphics[width=0.24\textwidth]{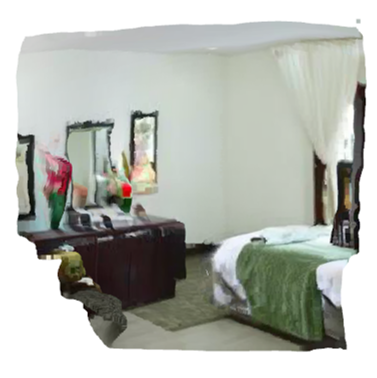}
      \includegraphics[width=0.24\textwidth]{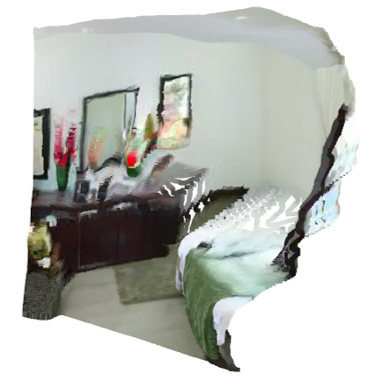}
      \includegraphics[width=0.24\textwidth]{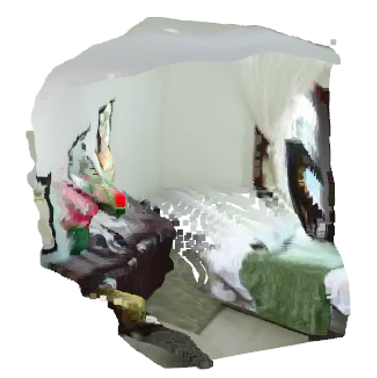}

      \caption{(a)}

\end{subfigure}
\hspace{1pt}
\begin{subfigure}{.49\textwidth}
\centering
\lineskip=0pt
      \includegraphics[width=0.24\textwidth]{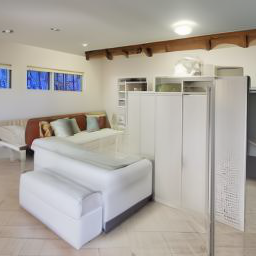}
      \includegraphics[width=0.24\textwidth]{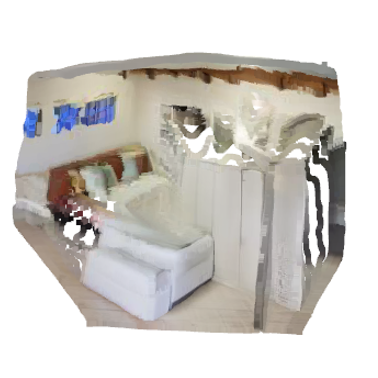}
      \includegraphics[width=0.24\textwidth]{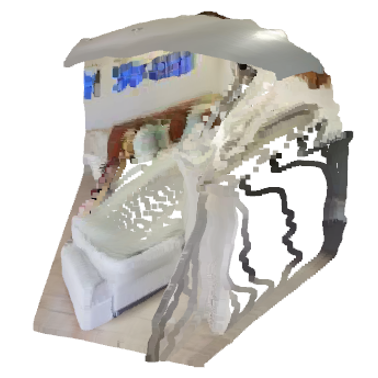}
      \includegraphics[width=0.24\textwidth]{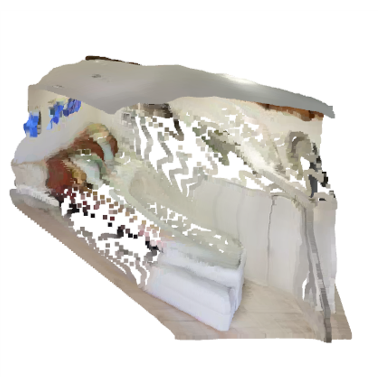}\\
      
      \includegraphics[width=0.24\textwidth]{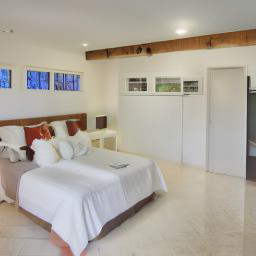}
      \includegraphics[width=0.24\textwidth]{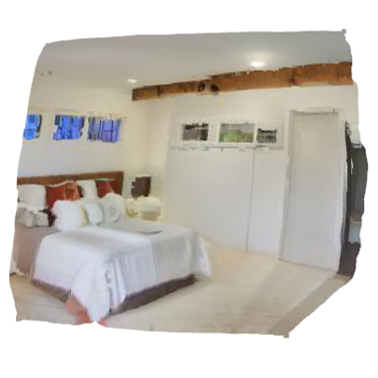}
      \includegraphics[width=0.24\textwidth]{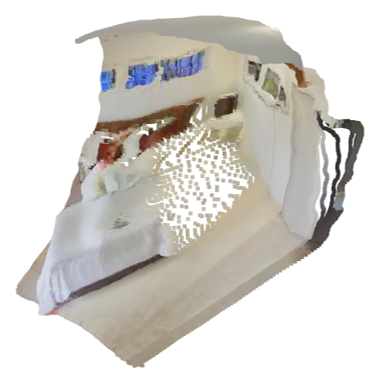}
      \includegraphics[width=0.24\textwidth]{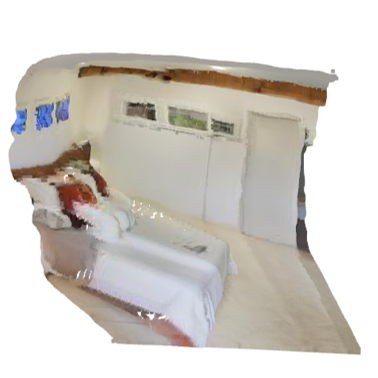}

      \caption{(b)}
\end{subfigure}
\vspace{-5pt}
\caption{\textbf{Visualization of point cloud representation obtained by depth information.} We compare the results generated from the baseline without (top) and with (bottom) our guidance by showing images and converting them into point cloud visualizations in four different views.}
\label{fig:pointcloud}
\vspace{-10pt}
\end{figure*}

%% file: latex/4_experiment.tex
\section{Experiments}

\subsection{Experimental Settings}

\paragraph{Datasets.}
In order to evaluate the performance of our proposed method, we conduct experiments on LSUN-Bedroom and LSUN-Church~\cite{yu2015lsun} for both depth estimation and image generation tasks. These datasets have been chosen to demonstrate depth-aware image synthesis in indoor and outdoor scenes, respectively. As there are no ground-truth depth labels available in the LSUN dataset, we generate pseudo-labels using a DPT~\cite{ranftl2021vision} pretrained on the NYU-Depth dataset~\cite{silberman2012indoor} and utilize them for training the depth estimator.
\vspace{-10pt}

\paragraph{Implementation details.}
We base our experiment on the label-efficient framework presented in \cite{baranchuk2022labelefficient}, where we modify the architecture of the pixel classifier to model a depth estimator. We use the pretrained weights from ADM~\cite{dhariwal2021diffusion} for LSUN-Bedroom, and for LSUN-Church, we build upon the publicly available repository of DDIM~\cite{song2021denoising} and Diffusers library~\cite{von-platen-etal-2022-diffusers}. We use Adam optimizer~\cite{kingma2014adam} to train the depth predictor MLP, and for the sampling of the diffusion model, we employ the DDIM sampler with a DDIM25 scheduler provided in the official repository of ADM~\cite{dhariwal2021diffusion} and DDIM. More detailed hyper-parameters and settings can be found in the appendix.
\vspace{-10pt}

\paragraph{Evaluation metrics for depth-awareness.}
Our primary contribution is the incorporation of geometric awareness in the image generation process, which is not reflected by the performance metrics commonly used in the image domain. Therefore, we introduce a novel performance metric for models that generate depth-guided images. First, we predict the depth maps of generated images using Dense Prediction Transformer (DPT)~\cite{ranftl2021vision}, a state-of-the-art monocular depth estimator. To measure the reality of the depth estimation map, we directly evaluate FID~\cite{shi20223d} with depth images and denote it dFID.
To make a fair comparison, we build the reference batch following~\cite{dhariwal2021diffusion} with the depth predictions of images from the dataset with DPT-Hybrid. 

\input{latex/TableBox/FID}

\subsection{Experimental Results}

\label{sec:result}
\paragraph{Depth prediction performance.}
First, to guide the diffusion sampling process using the depth predictor, we must validate its performance according to varying training dataset sizes, feature blocks, and timesteps. To provide guidance for synthesizing images at nearly all timesteps, we train the predictor for timesteps $t<800$. This is due to our observation that the depth predictions at $t\geq 800$ do not have any useful information. We evaluate the depth performance using the depth accuracy metric ($\delta<1.25$), and the results are shown in Fig.~\ref{fig:abl_depth}. Based on the results in Fig.~\ref{fig:abl_depth}~(a), we choose to use the middle feature blocks $\{l_n\}=\{2,4,5,6,7\}$, which show relatively high accuracy. In DCG, we also choose the feature maps by sorting the layer by accuracy, and the result is $S=\{2,4,5,6,7\}$ and $W=\{6\}$.
Next, Fig.~\ref{fig:abl_depth}~(b) illustrates the depth prediction accuracy with respect to the number of training images and evaluated timesteps. To minimize the sampling variance, We report the mean from a total of 10-fold training results. As the training data size increases, the performance of the model shows improvement accordingly. When trained with 1,000 images, the model achieves the highest accuracy. However, the performance increase compared to the 100-label setting is marginal, so we choose to use 100 images for training the depth predictor. \vspace{-10pt}

\paragraph{Quantitative results.}
We compare the results of evaluation metrics for LSUN-Bedroom between unguided ADM~\cite{dhariwal2021diffusion} and ADM with our guidance, and the results are shown in Tab.~\ref{tab:main_table}. The results indicate that compared to our baseline, ADM guided by DPG or DCG shows performance improvement in dFID, a metric used to evaluate performance in the depth domain. 
Furthermore, the performance of our guidance method is evaluated on the LSUN-Church~\cite{yu2015lsun} dataset utilizing the same metric, as shown in Tab.~\ref{tab:lsun_church}. Generated samples by using our method exhibit better performance in dFID. 

\input{latex/FigureBox/LsunChurch.tex}

\vspace{-10pt}
\paragraph{Qualitative results.}
In addition, we compare the result from ADM without our guidance and with our guidance method in Fig.~\ref{fig:qual_main}. We show both generated images and predicted depth maps using DPT, which demonstrates the effectiveness of our method of generating depth-aware images so that depth prediction models can predict the depth more robustly. Unlike guidance-free sampling, which is shown to generate geometrically implausible results, our method successfully generates depth-aware samples.
Moreover, we represent surface normal estimation~\cite{Bae_2021_ICCV} (Fig.~\ref{fig:qual_main}) and point cloud visualization (Fig.~\ref{fig:pointcloud}) to further show the effectiveness of our method in 3D scene understanding. The predictions made by our method show clearer boundaries and contain a higher level of details in scene geometry, as compared to the baseline.
We also provide qualitative comparisons on LSUN-Church between unguided samples and samples using our guidance method in Fig.~\ref{fig:qual_church}. Similar to the results observed on LSUN-Bedroom, we find that our guidance method can effectively preserve geometric characteristics.

\begin{figure}[t]
    \centering
    \begin{subfigure}{0.494\linewidth}
        \centering
        \includegraphics[width=1.0\linewidth]{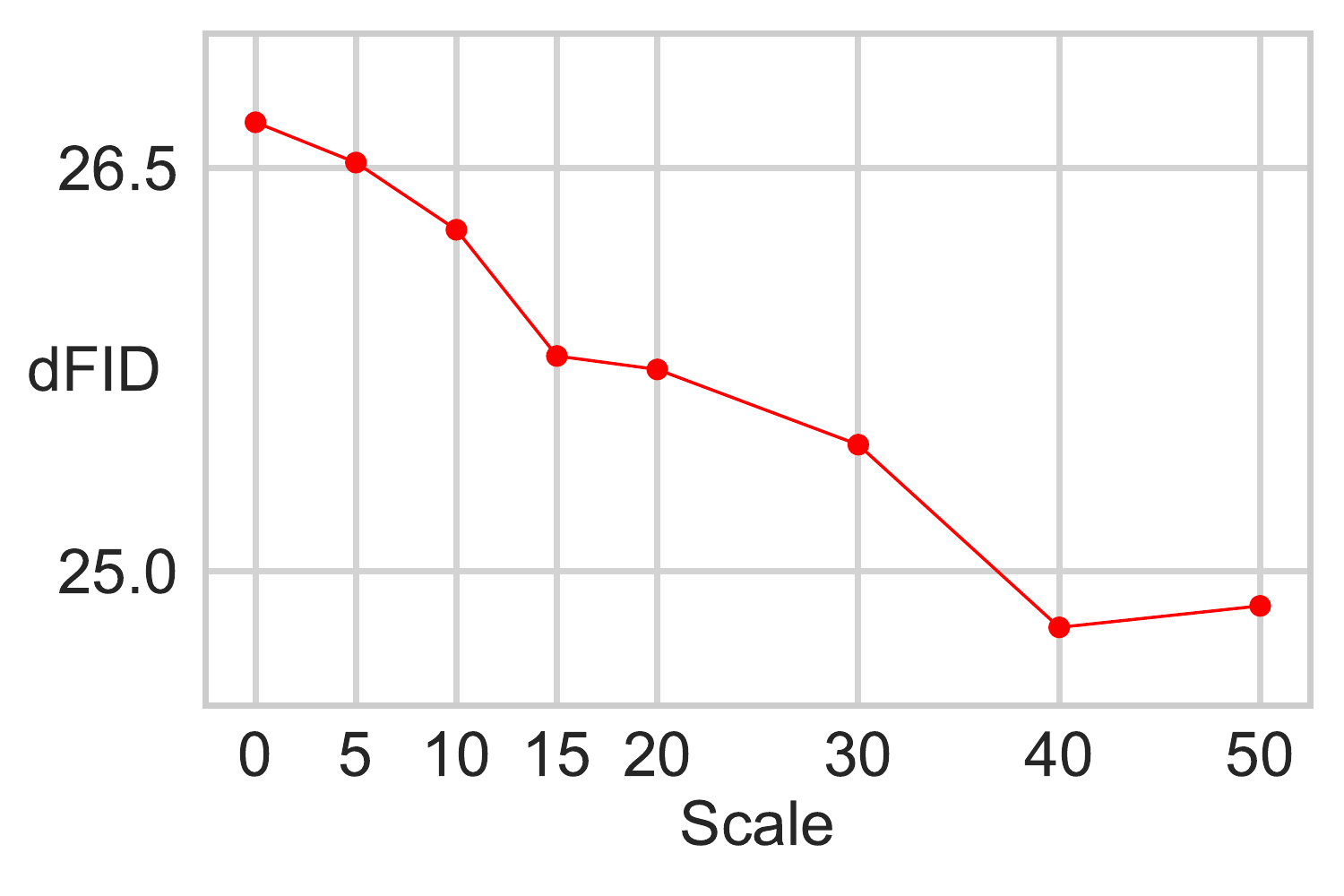}

        \caption{DCG}
        
    \end{subfigure}
    \begin{subfigure}{0.494\linewidth}
        \centering
        \includegraphics[width=1.0\linewidth]{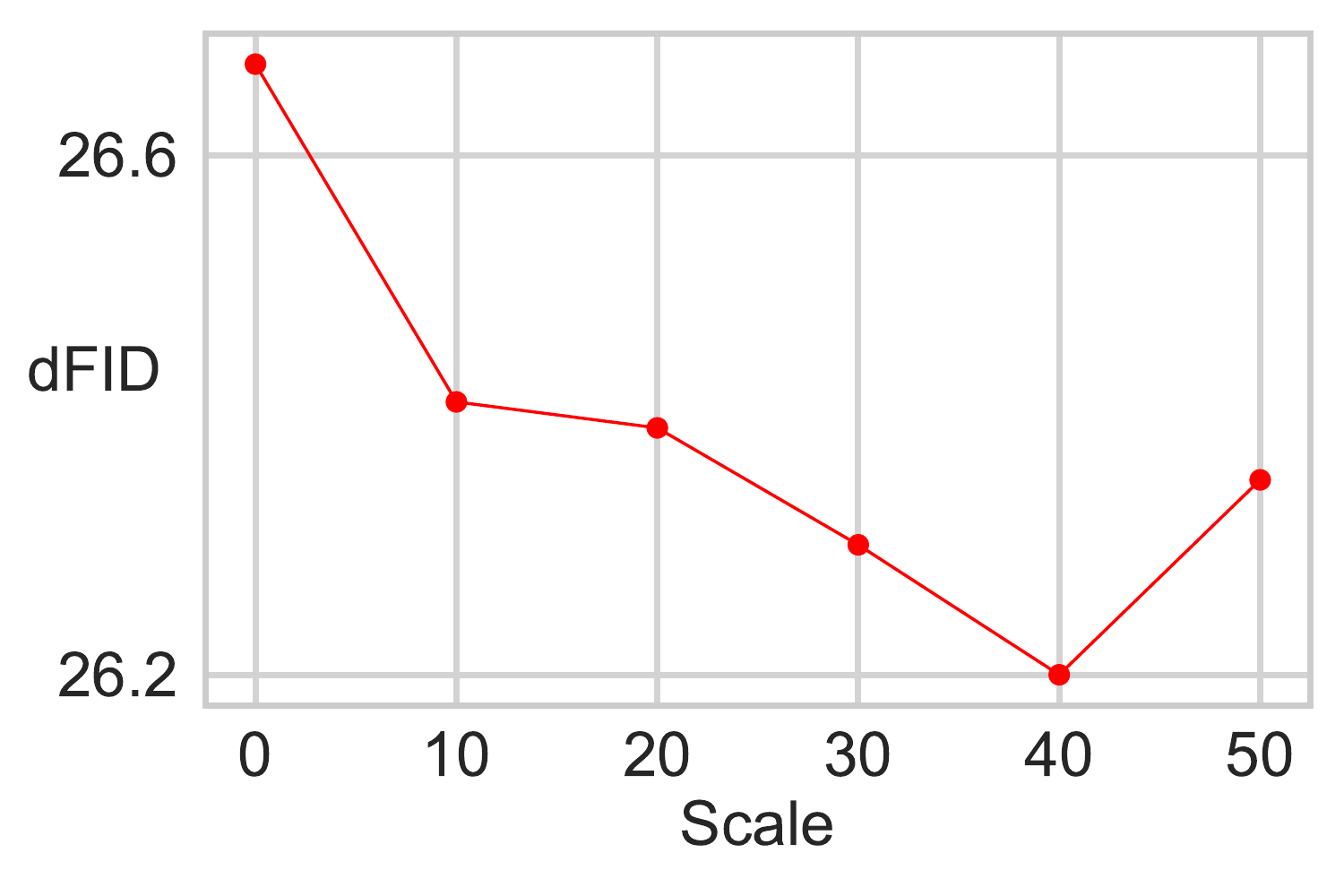}

        \caption{DPG}
    \end{subfigure}
\vspace{-5pt}
    \caption{\textbf{Comparison of dFID with respect to the guidance scales of DCG and DPG.}}
    \label{fig:abl-scale}

\end{figure}

\subsection{Ablation Study}
In this section, we analyze the effectiveness of different experimental settings and choices in our framework. To reduce the effect of random variation, we fix the random seed so that all variants see the same images in the same order. To compute the metrics, we perform all ablation studies on 1K generated samples.
\vspace{-10pt}

\input{latex/TableBox/prior.tex}

\input{latex/TableBox/downstream.tex}

\paragraph{Guidance scale.}
We study the relationship between the guidance scale and image quality in terms of dFID for each DCG and DPG. Fig.~\ref{fig:abl-scale} shows the tendency of the metrics versus scales of each guidance. As depicted in Fig.~\ref{fig:abl-scale}, DCG and DPG obtains the best results at $\omega_\mathrm{dc}=40$ and $\omega_\mathrm{dp}=40$ in dFID, respectively. Therefore, we treat this scale as default for the remaining part.
\vspace{-10pt}

\paragraph{Resolution of the prior network.}
In our second guidance method, DPG, we need to train a diffusion network to give a prior for condition image. But there are limitations in computation cost for the sampling process, so we choose the proper model size for the prior diffusion network. As our output depth map has a resolution of $64\times64$, we interpolate the depth map when fed to the prior network. We test three resolutions: $[32\times32, 64\times64, 128\times128]$ for the pretrained prior diffusion network, and the quantitative results are shown in Tab.~\ref{tab:prior}. The $128 \times 128$ outperforms the other resolution settings.

\subsection{Application for Monocular Depth Estimation}
To improve the effects of our generation as unlabeled data, we leverage the guided images and corresponding depth maps. We train the U-Net-based depth estimation network~\cite{godard2019digging} and evaluate the metrics with the NYU-Depth datasets~\cite{silberman2012indoor}. We compare the training results using reference data, unguided generated results, and our generated results. For the depth evaluation, we use accuracy under the threshold ($\delta<1.25$) and absolute relative error (AbsRel). Tab.~\ref{tab:downstream} shows that the images generated by DAG-based data are more helpful in training the depth predictor than the unguided samples set. 

%% file: latex/TableBox/FID.tex
\begin{table}[t]

\centering

\caption{\textbf{Quantitative Results on the LSUN-bedroom~\cite{yu2015lsun} dataset.} dFID denotes the FID score using the estimated depth image. The metrics are computed with 5,000 generated samples. The best result is in bold.}\label{tab:main_table}\vspace{-5pt}
\begin{tabular}{c|cc|cccccc} 
\toprule
Methods             & DPG  & DCG  & dFID ($\downarrow$) \\
\midrule
Baseline            & -     & -         & 15.71     \\ \midrule
\multirow{3}{*}{DAG}&\checkmark& -        & 14.18      \\ 
                    & -     &\checkmark     & 15.27      \\ 
                    &\checkmark&\checkmark     & \textbf{13.93}   \\ 
\bottomrule
\end{tabular}
\vspace{-10pt}
\end{table}

%% file: latex/FigureBox/LsunChurch.tex
\begin{table}[t]
\centering
\caption{\textbf{Quantitative results on the LSUN-church~\cite{yu2015lsun} dataset.} dFID denotes the FID score using the estimated depth image. The metrics are computed with 5,000 generated samples. The best result is in bold.}
\vspace{-5pt}
\begin{tabular}{c|cc|cc} 
\toprule
Methods             & DPG  & DCG  & dFID ($\downarrow$)  \\ 
\midrule
Baseline            & -     & -             & 17.69        \\ 
\midrule
\multirow{3}{*}{Ours}&\checkmark& -          & 17.43         \\ 
                    & -     &\checkmark     & 17.40        \\ 
                    &\checkmark&\checkmark  & \textbf{17.31}       \\ 
\bottomrule
\end{tabular}
\label{tab:lsun_church}
\end{table}

\begin{figure}[t]
\centering
\begin{subfigure}{1.0\linewidth}
\centering
\lineskip=0pt
      \includegraphics[width=0.245\linewidth]{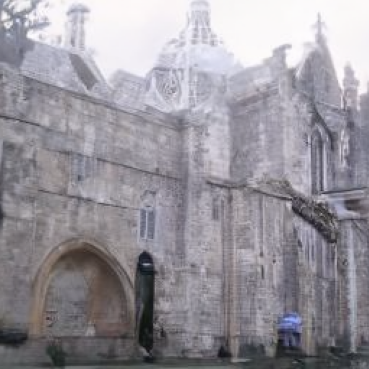}\hspace{-0.1em}
      \includegraphics[width=0.245\linewidth]{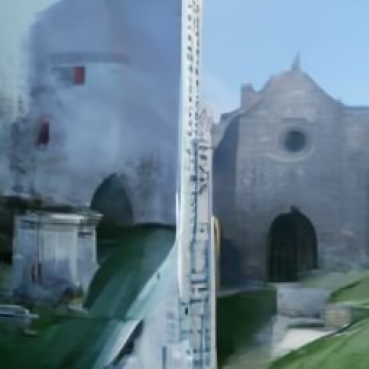}\hspace{-0.1em}
      \includegraphics[width=0.245\linewidth]{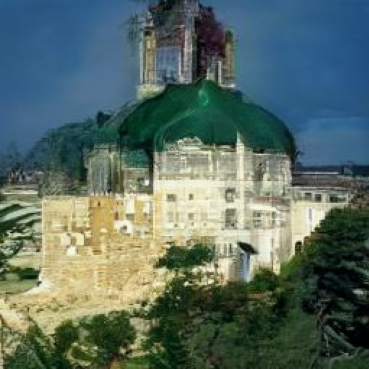}\hspace{-0.1em}
      \includegraphics[width=0.245\linewidth]{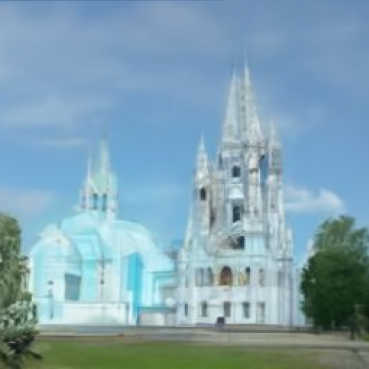}\hspace{-0.1em}

\end{subfigure}
\begin{subfigure}{1.0\linewidth}
\centering
\lineskip=0pt
      \includegraphics[width=0.245\linewidth]{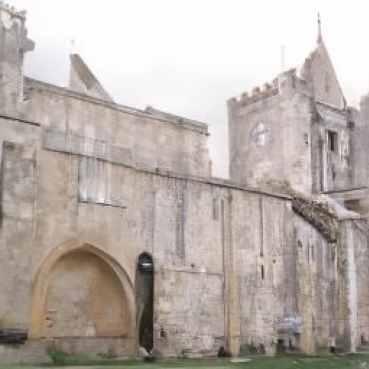}\hspace{-0.1em}
      \includegraphics[width=0.245\linewidth]{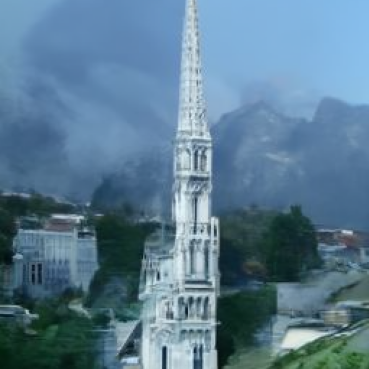}\hspace{-0.1em}
      \includegraphics[width=0.245\linewidth]{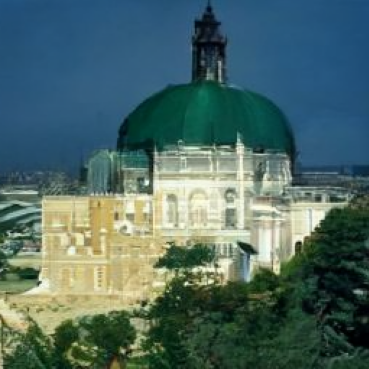}\hspace{-0.1em}
      \includegraphics[width=0.245\linewidth]{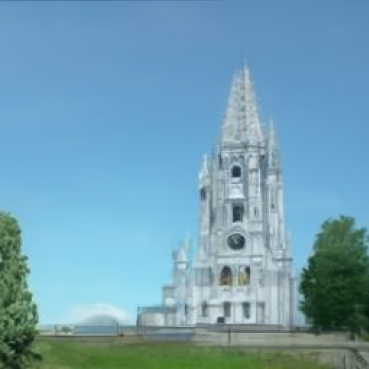}\hspace{-0.1em}

\end{subfigure}
\vspace{-10pt}
\caption{\textbf{Qualitative results on LSUN-church~\cite{yu2015lsun} dataset.} The first row is unguided samples from DDIM~\cite{song2021denoising}, and the second row is guided samples using our guidance method, DAG.}
\label{fig:qual_church}
\end{figure}

%% file: latex/TableBox/prior.tex
\begin{table}[t]
\centering
\caption{\textbf{Ablation study of the resolution of the depth prior network.}} 
\vspace{-5pt}
\begin{tabular}{c|cc}
\noalign{\smallskip}\noalign{\smallskip}\toprule
Method & dFID ($\downarrow$) \\
\midrule
Baseline & 26.77  \\
\midrule
$32\times32$ & 25.14  \\
$64\times64$ & 25.96  \\
$128\times128$ & \textbf{25.07}  \\
\bottomrule
\end{tabular}
\label{tab:prior}
\vspace{-5pt}
\end{table}

%% file: latex/TableBox/downstream.tex
\begin{table}[t]
\centering
\caption{\textbf{Application for monocular depth estimation.} We train the depth estimation model from scratch using U-Net~\cite{ronneberger2015u} based backbone network with our synthesized data.} 
\vspace{-5pt}
\begin{tabular}{c|cc}
\noalign{\smallskip}\noalign{\smallskip}\toprule
Method & $\delta>1.25$ ($\uparrow$) & AbsRel. ($\downarrow$) \\
\midrule
Supervised & 79.06 & 0.144 \\
\midrule
Unguided data & 72.55 & 0.185 \\
DAG-based data & \textbf{77.54} & \textbf{0.151} \\
\bottomrule
\end{tabular}

\vspace{-10pt}
\label{tab:downstream}
\end{table}

%% file: latex/5_conclusion.tex
\section{Conclusion}
In this paper, we propose a label-efficient method for predicting depth maps of images generated by the reverse process of diffusion model using internal representations. We also introduce a novel guidance scheme to guide the image to have a plausible depth map. In addition, to address the problem that existing measurement methods fail to capture the depth feasibility of generated images, we present a new evaluation metric that effectively represents this depth awareness using pretrained depth estimation networks.

%% file: supple/supple.tex
\onecolumn


\input{supple/1.Metric.tex}
\input{supple/2.Qual.tex}

\input{supple/3.Pointcloud.tex}
\input{supple/4.Surface_normal.tex}

%% file: supple/1.Metric.tex
\section{Further Details}
\label{sec:detail}
\subsection{Implementation Details}
\paragraph{Hyperparameter Settings.}
We report hyperparameters utilized during the training of the depth estimator and sampling process in Tab.~\ref{tab:hparam}. For two datasets, LSUN-Bedroom and LSUN-Church~\cite{yu2015lsun}, the hyperparameters employed for the depth estimator training are largely consistent. Note that we use two different U-Net implementations, namely ADM~\cite{dhariwal2021diffusion} and DDIM~\cite{song2021denoising}, due to the existence of pretrained weights, thus leading to slight variations in the selection of the feature extraction block.

\begin{table*}[h]
    \centering
    \small
    \caption{\textbf{Detailed hyperparameter settings.} Training denotes hyperparameters used at the training depth estimator, and sampling denotes hyperparameters used at sampling diffusion models using DAG.}
    \vspace{-5pt}
    \begin{tabular}{c|c|cc}
    \toprule
       & Hyperparameters                            & LSUN-Bedroom~\cite{yu2015lsun} & LSUN-Church~\cite{yu2015lsun}\\ \midrule
\multirow{6}{*}{Training}       & learning rate     & $1e-5$        & $1e-5$ \\
                                & optimizer         & Adam          & Adam \\ 
                                & $\mathbf{g}^{W}$  & $[\mathbf{f}_t(6)]$ & $[\mathbf{f}_t(17)]$   \\
                                & $\mathbf{g}^{S}$  & $[\mathbf{f}_t(2); \mathbf{f}_t(4); \mathbf{f}_t(5); \mathbf{f}_t(6); \mathbf{f}_t(7)]$ & $[\mathbf{f}_t(3); \mathbf{f}_t(5); \mathbf{f}_t(7); \mathbf{f}_t(9); \mathbf{f}_t(13); \mathbf{f}_t(17)]$   \\ 
                                & Timestep Sampler  & Uniform       & Uniform \\ 
                                & Time embedding dimension & 256 & 256 \\ \midrule
\multirow{5}{*}{Sampling}       & Timestep          & 25            & 25   \\
                                & Schedular         & DDIM          & DDIM  \\
                                & $\tau$            & Uniform          & Uniform  \\
                                & $\omega_{dc}$     & 40.0          & 10.0  \\
                                & $\omega_{dp}$     & 40.0          & 50.0  \\ \bottomrule
    \end{tabular}
    \label{tab:hparam}
    \vspace{-10pt}
\end{table*}

\subsection{PyTorch-like Pseudo Code.}
\paragraph{Guidance Method.} We provide a detailed PyTorch-like pseudo-code of our guidance method in Algorithm~\ref{alg:guide}. \vspace{-10pt}

\paragraph{Network Architecture for Depth Estimator.}
We use a 3-layer MLP that receives concatenated feature vector of a pixel to predict the depth map. Following the diffusion network, we apply the SiLU activation function. Detailed architecture is shown in Fig.~2 of the main paper and Algorithm~\ref{alg:est} of this appendix.
Also, we add a time-embedding module to this depth estimator MLP following the implementation of U-Net of ADM~\cite{dhariwal2021diffusion}.

\begin{figure*}[t]
\begin{minipage}[c]{\linewidth}
\begin{algorithm}[H]
\small
\caption{\small Pseudo Code of Our Depth-Aware Guidance}
\label{alg:guide}
\definecolor{codeblue}{rgb}{0.25,0.5,0.5}
\definecolor{codegreen}{rgb}{0,0.6,0}
\definecolor{codekw}{rgb}{0.85, 0.18, 0.50}
\lstset{
  backgroundcolor=\color{white},
  basicstyle=\linespread{1.0}\fontsize{9pt}{9pt}\ttfamily\selectfont,
  columns=fullflexible,
  breaklines=true,
  captionpos=b,
  commentstyle=\fontsize{9pt}{9pt}\color{codegreen},
  keywordstyle=\fontsize{9pt}{9pt}\color{codekw},
  escapechar={|}, 
  xleftmargin=0.0in,
  xrightmargin=0.0in
}
\begin{lstlisting}[language=python]
def cond_fn(img, t, y=None):
    """
    dc: scale of depth consistency guidance
    dp: scale of depth prior guidance
    """
    img = img.detach().requires_grad_(True)

    # feature extraction
    strong_feature, weak_feature = feature_extractor(img, t)

    # depth consistency loss
    strong_depth = strong_depth_predictor(strong_feature, t)
    weak_depth = weak_depth_predictor(weak_feature, t)
    dcg_loss = F.l1_loss(weak_depth, strong_depth.detach())

    # depth prior loss
    noise = torch.randn_like(strong_depth)
    t_rand = torch.randint(0, 25)

    strong_depth_t = diffusion.q_sample(strong_depth, t_rand, noise=noise)
    strong_eps = prior_diffusion_model(strong_depth_t, t_rand)
    dpg_loss = torch.mean((strong_eps - noise) ** 2)

    loss = dp * dpg_loss + dc * dcg_loss
    grad = torch.autograd.grad(loss, img, allow_unused=True)[0]

    return -1.0 * grad
\end{lstlisting}
\end{algorithm}
\end{minipage}
\hfill
\begin{minipage}[c]{\linewidth}
\begin{algorithm}[H]
\small
\caption{\small Pseudo Code of Our Label-efficient Depth Estimator }
\begin{multicols}{2}
\label{alg:est}
\definecolor{codeblue}{rgb}{0.25,0.5,0.5}
\definecolor{codegreen}{rgb}{0,0.6,0}
\definecolor{codekw}{rgb}{0.85, 0.18, 0.50}
\lstset{
  backgroundcolor=\color{white},
  basicstyle=\linespread{1.0}\fontsize{9pt}{9pt}\ttfamily\selectfont,
  columns=fullflexible,
  breaklines=true,
  captionpos=b,
  commentstyle=\fontsize{9pt}{9pt}\color{codegreen},
  keywordstyle=\fontsize{9pt}{9pt}\color{codekw},
  escapechar={|}, 
  xleftmargin=0.0in,
  xrightmargin=0.2in,
}
\begin{lstlisting}[language=python]
class MLPBlock(nn.Module):
    def __init__(self, in_dim, out_dim, time_dim):
        super().__init__()
        self.linear = nn.Linear(in_dim, out_dim)
        self.is_last = True if out_dim == 1 else False
        
        if not self.is_last:
            self.act = nn.SiLU()
            self.norm = nn.BatchNorm1d(out_dim)
            self.time_proj = nn.Linear(time_dim, out_dim)
        else:
            self.act = nn.Sigmoid()
            self.norm = nn.Identity()
            self.time_proj = None

    def forward(self, x, time_emb=None):
        x = self.linear(x)
        x = self.act(x)
        x = self.norm(x)
        if not self.is_last:
            x = x + self.time_proj(time_emb)
        return x
\end{lstlisting}
\columnbreak
\begin{lstlisting}[language=python]
class DepthPredictor(nn.Module):
    def __init__(self, dim):
        super().__init__()
        self.layers = nn.ModuleList([
            MLPBlock(dim, 256, time_dim=256),
            MLPBlock(256, 128, time_dim=256),
            MLPBlock(128, 1, time_dim=256),
        ])
        self.max_depth = 10.0

    def forward(self, x, timesteps=None):
        timesteps = timestep_embedding(timesteps, 256)
        
        for layer in self.layers:
            x = layer(x, timesteps)
            
        return x * self.max_depth
\end{lstlisting}
\end{multicols}
\end{algorithm}
\end{minipage}
\end{figure*}

\subsection{Proposed Metric}
\input{supple/figure/motivation.tex}

We propose a new metric, called dFID, which can evaluate the depth-awareness, and we describe the details of the proposed metrics in this section. \vspace{-10pt}

\paragraph{Depth FID.}
To compute the FID of the depth domain, we first estimate the depth map using the NYU-Depth~\cite{silberman2012indoor} pretrained depth estimator DPT-Hybrid~\cite{ranftl2021vision}, which is available in the official repository. Then, identical to the original FID, we compute the Fréchet distance of embeddings collected from depth images using the Inception v3 model~\cite{szegedy2015rethinking}. 
We show examples in Tab.~\ref{tab:motivation} where FID is good but the geometric realism is limited. It demonstrates that the FID is inappropriate for measuring geometrical awareness. To address this problem, we propose dFID. DAG, our proposed guidance method, improves both the visual quality of geometric realism and the dFID metric. However, in terms of FID, we can find that FID does not adequately capture the geometric plausibility.


\vspace{-10pt}


%% file: supple/figure/motivation.tex
\begin{table*}[h]
\centering
\caption{\textbf{Comparison of generated samples from ADM.} The images in the left two columns are examples of samples from ADM without our guidance method, DAG, and the images in the right two columns are examples from ADM using our method. 
}
\vspace{-5pt}
\begin{tabular}{c|cc|cc} 
\toprule
Samples  &                                         \raisebox{-.45\height}{\includegraphics[width=0.16\textwidth]{figures/figure1/image/2_no.png}} &                \raisebox{-.45\height}{\includegraphics[width=0.16\textwidth]{figures/figure1/image/4_no.png}} &                 \raisebox{-.45\height}{\includegraphics[width=0.16\textwidth]{figures/figure1/image/2_g.png}}  &                 \raisebox{-.45\height}{\includegraphics[width=0.16\textwidth]{figures/figure1/image/4_g.png}}\\ 
\midrule
DAG & \multicolumn{2}{c|}{\xmark} & \multicolumn{2}{c}{\checkmark} \\
\midrule
dFID ($\downarrow$) & \multicolumn{2}{c|}{15.71} & \multicolumn{2}{c}{\textbf{13.93}}\\ 
FID ($\downarrow$)  & \multicolumn{2}{c|}{\textbf{6.72}}  &  \multicolumn{2}{c}{7.59} \\  
\bottomrule
\end{tabular}
\label{tab:motivation}
\end{table*}

%% file: supple/2.Qual.tex
\section{More Visualizations}
\label{sec:qual}
\begin{figure*}[p]
\centering
\lineskip=0pt
\begin{subfigure}{1.0\linewidth}
\centering
\lineskip=0pt
      \includegraphics[width=0.15\linewidth]{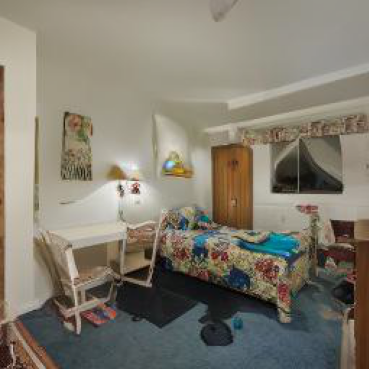}\hspace{-0.1em}
      \includegraphics[width=0.15\linewidth]{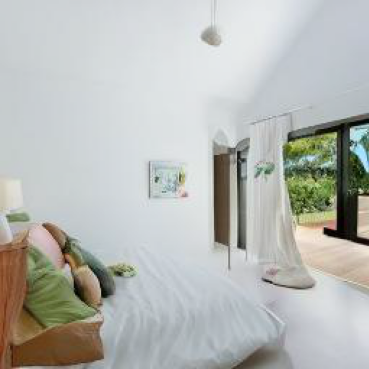}\hspace{-0.1em}
      \includegraphics[width=0.15\linewidth]{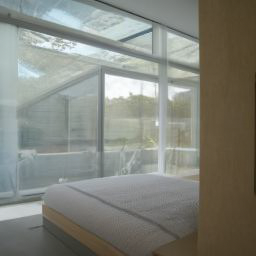}\hspace{-0.1em}
      \includegraphics[width=0.15\linewidth]{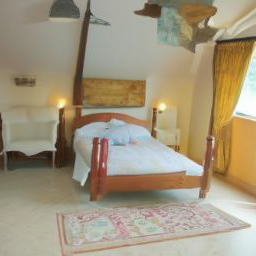}\hspace{-0.1em}
      \includegraphics[width=0.15\linewidth]{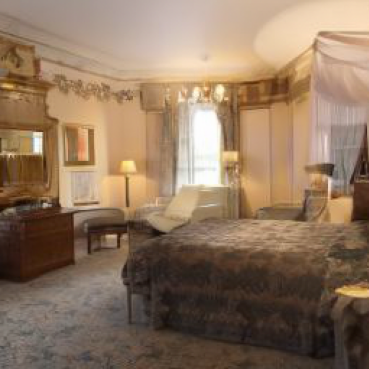}\hspace{-0.1em}
      \includegraphics[width=0.15\linewidth]{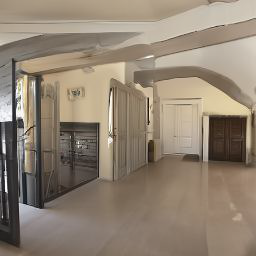}\hspace{-0.1em}

\end{subfigure}
\begin{subfigure}{1.0\textwidth}
\centering
\lineskip=0pt
      \includegraphics[width=0.15\linewidth]{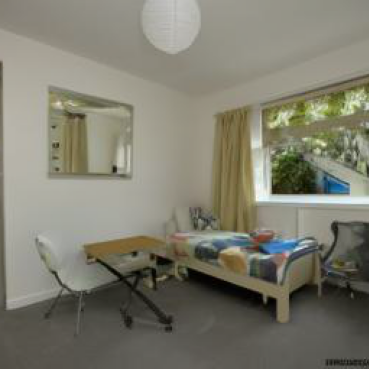}\hspace{-0.1em}
      \includegraphics[width=0.15\linewidth]{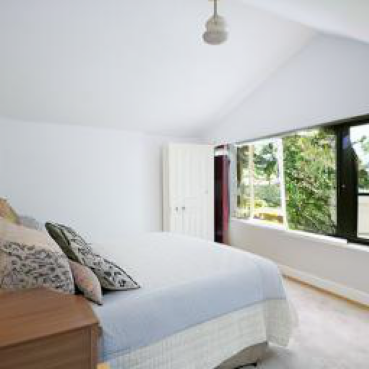}\hspace{-0.1em}
      \includegraphics[width=0.15\linewidth]{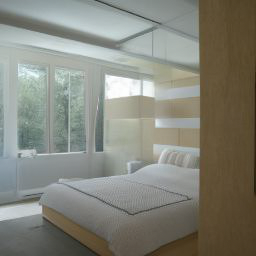}\hspace{-0.1em}
      \includegraphics[width=0.15\linewidth]{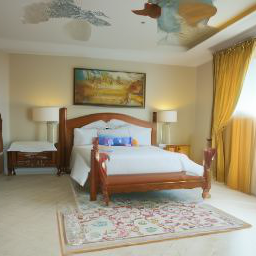}\hspace{-0.1em} 
      \includegraphics[width=0.15\linewidth]{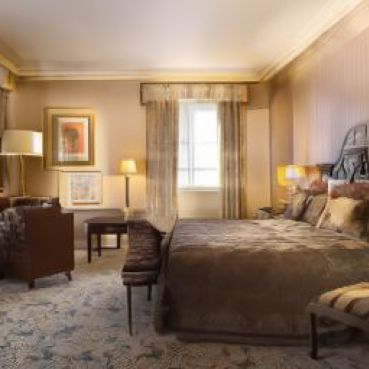}\hspace{-0.1em}
      \includegraphics[width=0.15\linewidth]{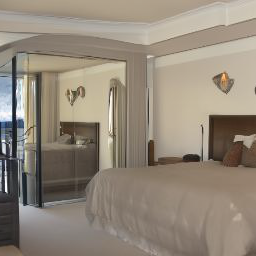}\hspace{-0.1em}

\end{subfigure}
\begin{subfigure}{1.0\textwidth}
\centering
\lineskip=0pt
      \includegraphics[width=0.15\linewidth]{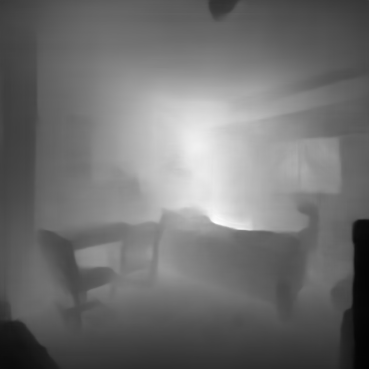}\hspace{-0.1em}
      \includegraphics[width=0.15\linewidth]{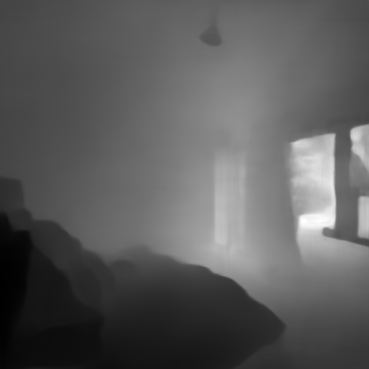}\hspace{-0.1em}
      \includegraphics[width=0.15\linewidth]{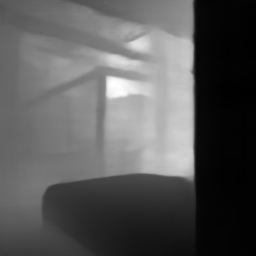}\hspace{-0.1em}
      \includegraphics[width=0.15\linewidth]{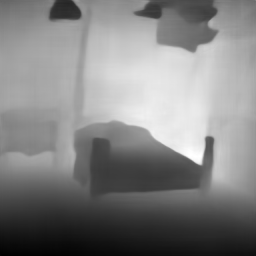}\hspace{-0.1em}
      \includegraphics[width=0.15\linewidth]{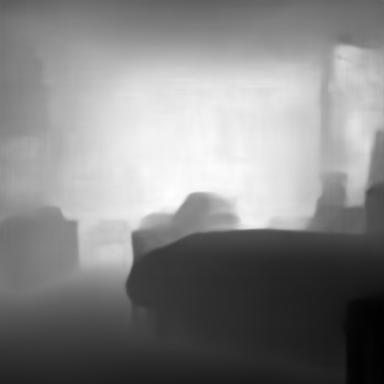}\hspace{-0.1em}
      \includegraphics[width=0.15\linewidth]{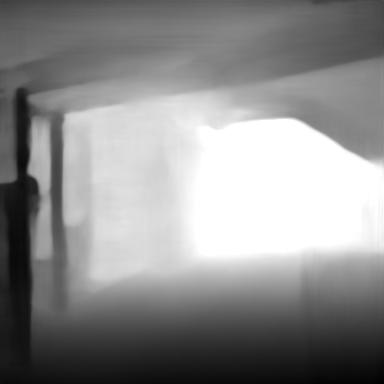}\hspace{-0.1em}

\end{subfigure}
\begin{subfigure}{1.0\textwidth}
\centering
\lineskip=0pt
      \includegraphics[width=0.15\linewidth]{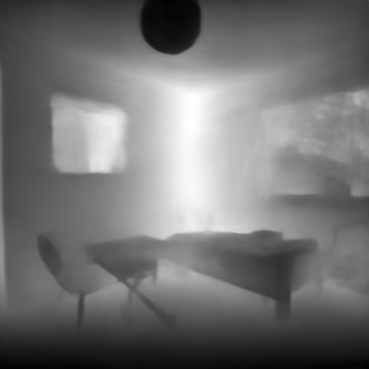}\hspace{-0.1em}
      \includegraphics[width=0.15\linewidth]{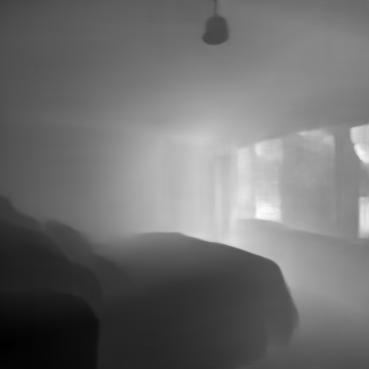}\hspace{-0.1em}
      \includegraphics[width=0.15\linewidth]{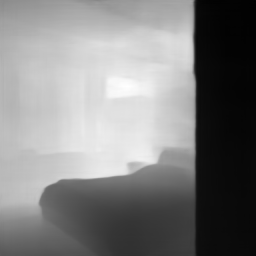}\hspace{-0.1em}
      \includegraphics[width=0.15\linewidth]{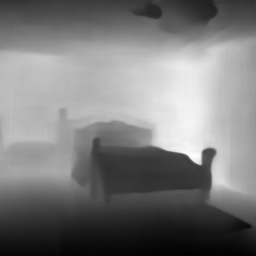}\hspace{-0.1em}
      \includegraphics[width=0.15\linewidth]{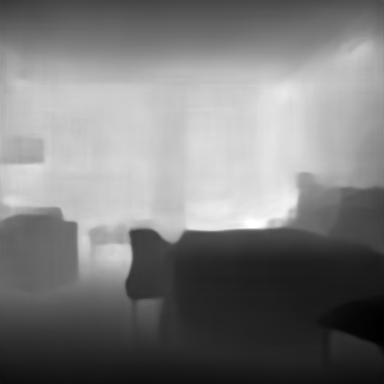}\hspace{-0.1em}
      \includegraphics[width=0.15\linewidth]{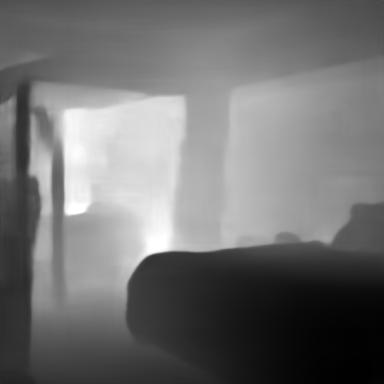}\hspace{-0.1em}
\end{subfigure}

\vspace{5pt}

\begin{subfigure}{1.0\textwidth}
\centering
\lineskip=0pt
      \includegraphics[width=0.15\linewidth]{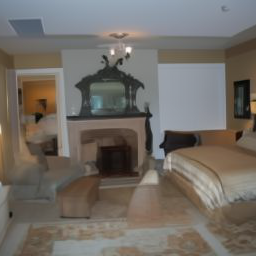}\hspace{-0.1em}
      \includegraphics[width=0.15\linewidth]{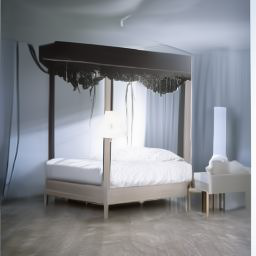}\hspace{-0.1em}
      \includegraphics[width=0.15\linewidth]{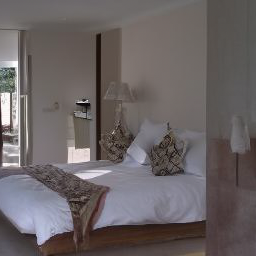}\hspace{-0.1em}
      \includegraphics[width=0.15\linewidth]{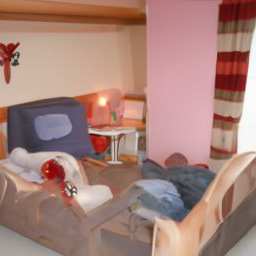}\hspace{-0.1em}
      \includegraphics[width=0.15\linewidth]{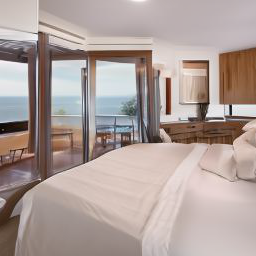}\hspace{-0.1em}
      \includegraphics[width=0.15\linewidth]{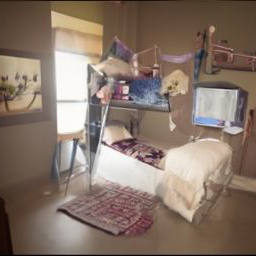}\hspace{-0.1em}

\end{subfigure}
\begin{subfigure}{1.0\textwidth}
\centering
\lineskip=0pt
      \includegraphics[width=0.15\linewidth]{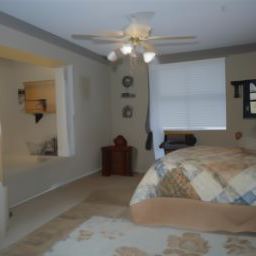}\hspace{-0.1em}
      \includegraphics[width=0.15\linewidth]{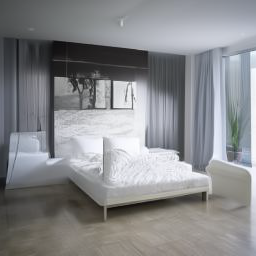}\hspace{-0.1em}
      \includegraphics[width=0.15\linewidth]{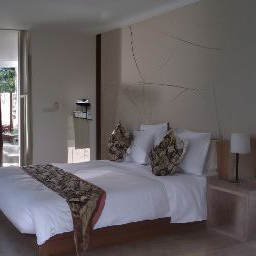}\hspace{-0.1em}
      \includegraphics[width=0.15\linewidth]{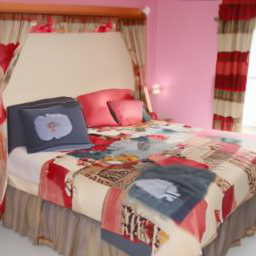}\hspace{-0.1em}
      \includegraphics[width=0.15\linewidth]{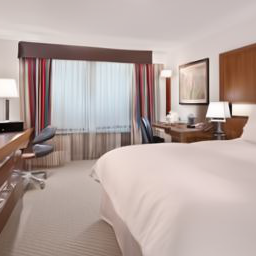}\hspace{-0.1em}
      \includegraphics[width=0.15\linewidth]{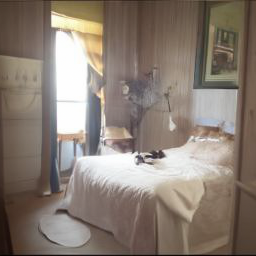}\hspace{-0.1em}

\end{subfigure}

\begin{subfigure}{1.0\textwidth}
\centering
\lineskip=0pt
      \includegraphics[width=0.15\linewidth]{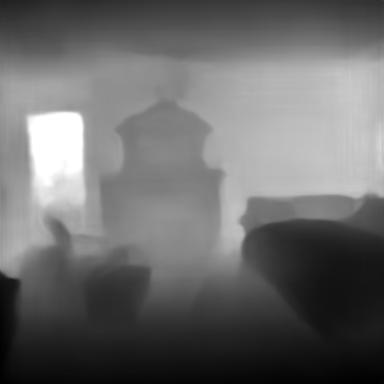}\hspace{-0.1em}
      \includegraphics[width=0.15\linewidth]{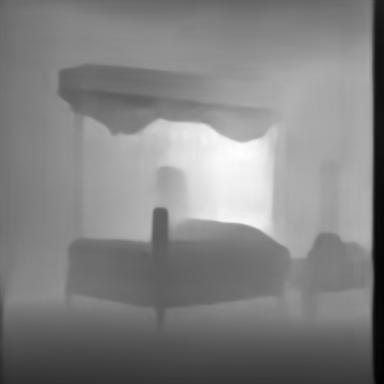}\hspace{-0.1em}
      \includegraphics[width=0.15\linewidth]{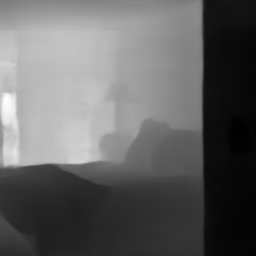}\hspace{-0.1em}
      \includegraphics[width=0.15\linewidth]{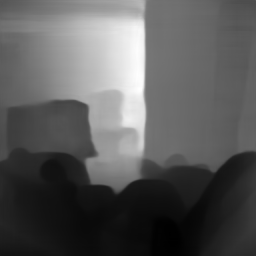}\hspace{-0.1em}
      \includegraphics[width=0.15\linewidth]{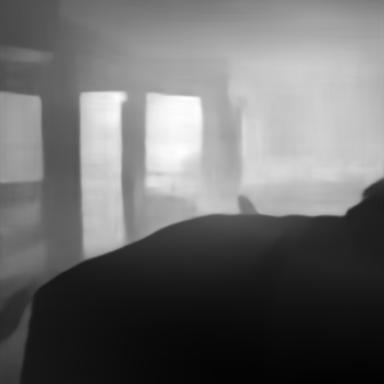}\hspace{-0.1em}
      \includegraphics[width=0.15\linewidth]{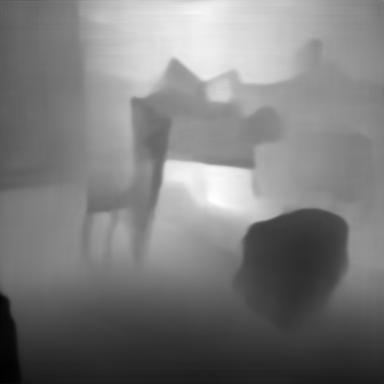}\hspace{-0.1em}

\end{subfigure}
\begin{subfigure}{1.0\textwidth}
\centering
\lineskip=0pt
      \includegraphics[width=0.15\linewidth]{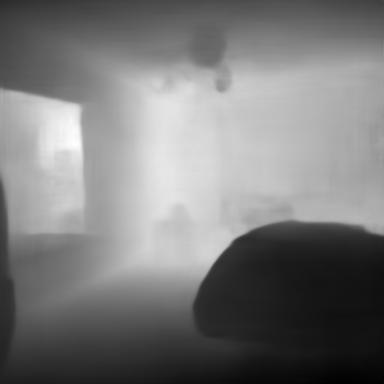}\hspace{-0.1em}
      \includegraphics[width=0.15\linewidth]{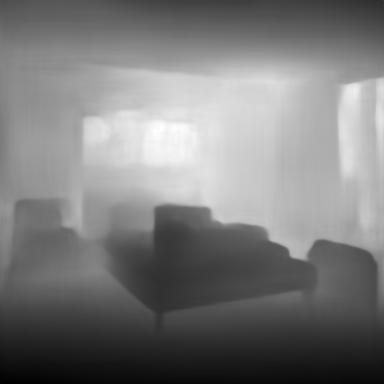}\hspace{-0.1em}
      \includegraphics[width=0.15\linewidth]{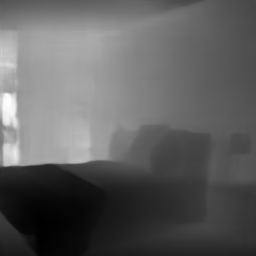}\hspace{-0.1em}
      \includegraphics[width=0.15\linewidth]{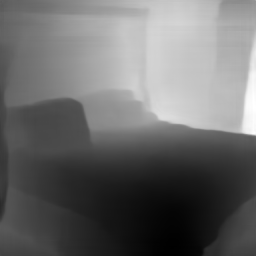}\hspace{-0.1em}
      \includegraphics[width=0.15\linewidth]{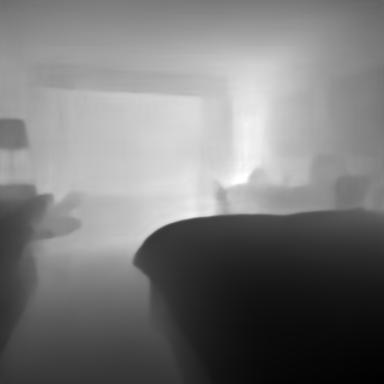}\hspace{-0.1em}
      \includegraphics[width=0.15\linewidth]{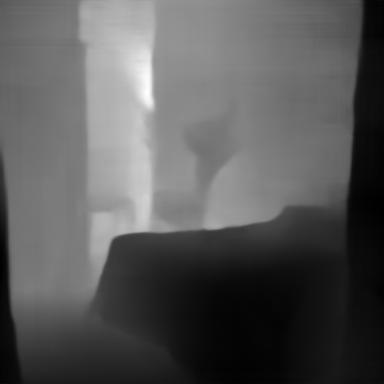}\hspace{-0.1em}

\end{subfigure}

\vspace{-5pt}
\caption{\textbf{Qualitative comparison on LSUN bedroom datasets~\cite{yu2015lsun}.} We show the results of unguided and depth-guided samples (first two rows) and their corresponding depths (last two rows). }
\label{fig:qual_supp}

\end{figure*}

Furthermore, we show additional qualitative results in Fig.~\ref{fig:qual_supp} comparing the performance of the baseline and our proposed guidance approach. The images generated using our guidance approach exhibit reduced perspective ambiguity and a more accurate portrayal of the layout when compared to the baseline. Additionally, through visualizations of the related depth maps, it is evident that the diffusion models guided by our approach produce more geometrically plausible outcomes than those generated without guidance.

\subsection{Point Cloud Visualization}
We provide supplementary results in Fig.~\ref{fig:pcd} that demonstrate the point cloud derived from the images, as well as its related depth, to highlight the depth awareness and geometric plausibility of our proposed guidance approach. To provide a comprehensive evaluation, we have generated point clouds from multiple perspectives, such as front, left, right, and upward, using the Open3D library~\cite{Zhou2018}, in order to compare the 3D-represented images. The point cloud representations of the guided samples are observed to be more consistent and realistic when viewed from the same perspective, and they exhibit robust geometric consistency.

\subsection{Surface Normal Visualization}
Fig.~\ref{fig:normal} also presents the results of surface normal estimation for both the baseline method and our proposed methodology. We employ surface normals to evaluate the geometrical rationality of the generated images as they are a crucial feature of geometric surfaces. Furthermore, by computing the probability distribution of the per-pixel surface normal, we estimate the aleatoric uncertainty~\cite{Bae_2021_ICCV}. It can be observed that when compared to the baseline, our method's predictions exhibit a more defined layout and bounds, resulting in lower uncertainty along the borders of walls.

\section{Limitations and Future Works}
\label{sec:lim}
In this study, we focus on the sampling strategy that leads to the generation of images with higher depth awareness. This strategy enables the use of generated images with more reliable and precise geometric properties in a variety of downstream tasks. As our work represents the first attempt to enable depth awareness in image synthesis using diffusion models, it is challenging to compare its performance to previous research. we have employed pretrained diffusion models in this study due to the high computational cost of the training, and the data selection was also limited. In future research, we plan to train the diffusion model using indoor and outdoor scene datasets to directly demonstrate and improve the effectiveness of our guiding technique. 
Additionally, while we have employed the diffusion U-Net as part of the depth estimator, this approach increases the cost of the guiding process through the gradient computation demands of backpropagation through the whole U-Net. An alternative approach to address this challenge would be to train the diffusion network using depth map conditioning and sampling with classifier-free guidance, however, this would require a significant amount of depth-image pairs and computational resources for joint training, compared to our current method.

\newpage

%% file: supple/3.Pointcloud.tex
\begin{figure*}[p]
\centering
\captionsetup[subfigure]{labelformat=empty}
\captionsetup[subfigure]{font=small}
\begin{tabular}{lcc}
\small
\centering
(a) &
\begin{minipage}[c]{0.19\textwidth}
\begin{subfigure}{1.0\textwidth}
  \centering
  \includegraphics[width=1.0\textwidth]{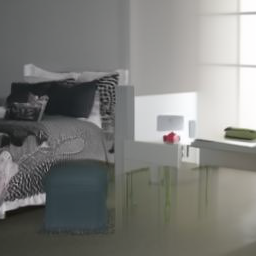}
\end{subfigure}
\end{minipage}
&
\begin{minipage}[c]{0.76\textwidth}
\begin{subfigure}{0.245\textwidth}
  \centering
  \includegraphics[width=1\linewidth]{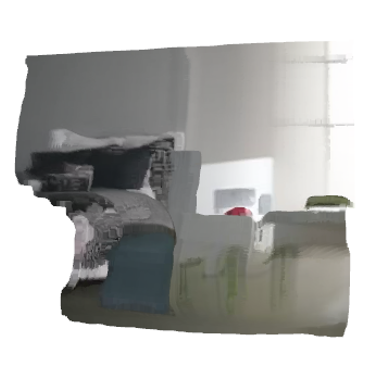}
\end{subfigure}
\begin{subfigure}{0.245\textwidth}
  \centering
  \includegraphics[width=1\linewidth]{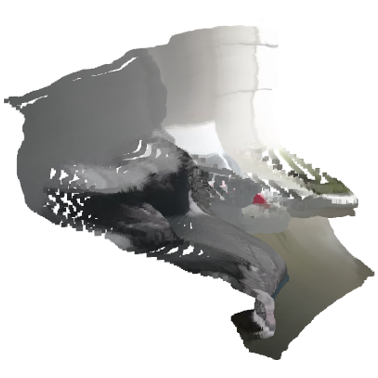}
\end{subfigure}
\begin{subfigure}{0.245\textwidth}
  \centering
  \includegraphics[width=1\linewidth]{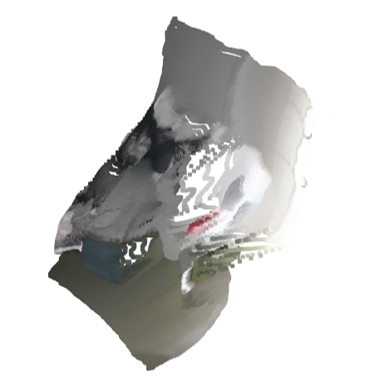}
\end{subfigure}
\begin{subfigure}{0.245\textwidth}
  \centering
  \includegraphics[width=1\linewidth]{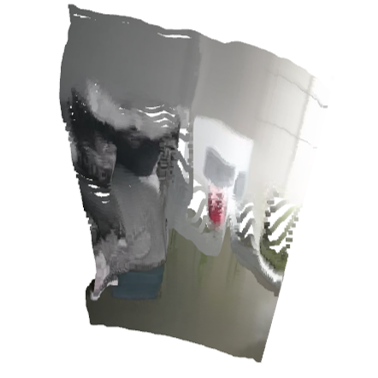}
\end{subfigure}
\end{minipage}
\vspace{3pt} \\ 

(b) &
\begin{minipage}[c]{0.19\textwidth}
\begin{subfigure}{1.0\textwidth}
  \centering
  \includegraphics[width=1\textwidth]{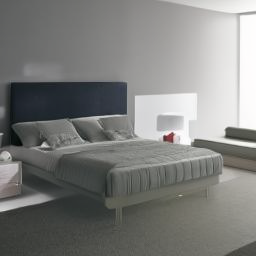}
\end{subfigure}
\end{minipage}
&
\begin{minipage}[c]{0.76\textwidth}
\begin{subfigure}{0.245\textwidth}
  \centering
  \includegraphics[width=1\linewidth]{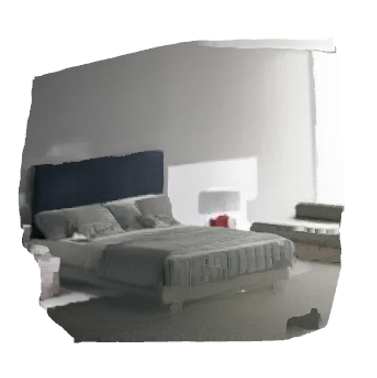}
\end{subfigure}
\begin{subfigure}{0.245\textwidth}
  \centering
  \includegraphics[width=1\linewidth]{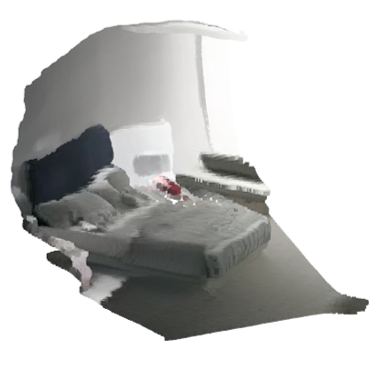}
\end{subfigure}
\begin{subfigure}{0.245\textwidth}
  \centering
  \includegraphics[width=1\linewidth]{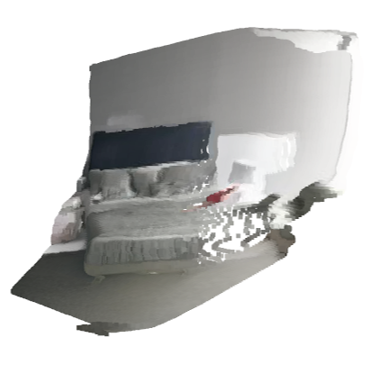}
\end{subfigure}
\begin{subfigure}{0.245\textwidth}
  \centering
  \includegraphics[width=1\linewidth]{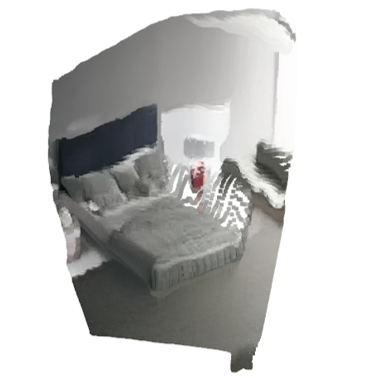}
\end{subfigure}
\end{minipage}
\vspace{3pt} \\

(c) &
\begin{minipage}[c]{0.19\textwidth}
\begin{subfigure}{1.0\textwidth}
  \centering
  \includegraphics[width=1\textwidth]{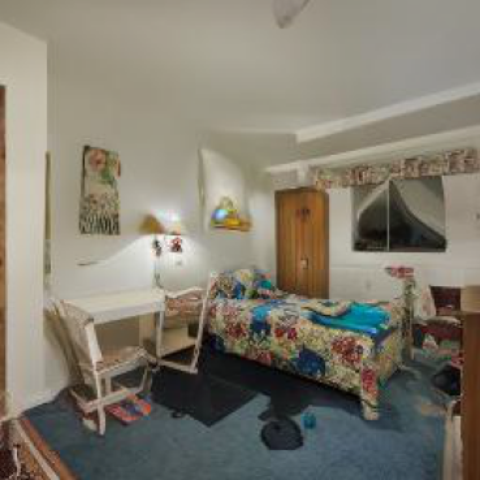}
\end{subfigure}
\end{minipage}
&
\begin{minipage}[c]{0.76\textwidth}
\begin{subfigure}{0.245\textwidth}
  \centering
  \includegraphics[width=1\linewidth]{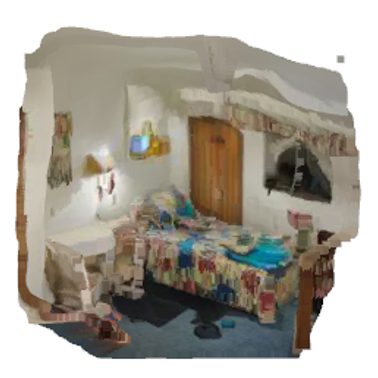}
\end{subfigure}
\begin{subfigure}{0.245\textwidth}
  \centering
  \includegraphics[width=1\linewidth]{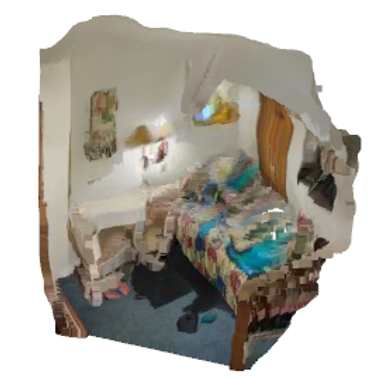}
\end{subfigure}
\begin{subfigure}{0.245\textwidth}
  \centering
  \includegraphics[width=1\linewidth]{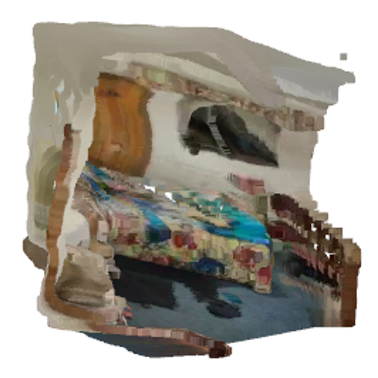}
\end{subfigure}
\begin{subfigure}{0.245\textwidth}
  \centering
  \includegraphics[width=1\linewidth]{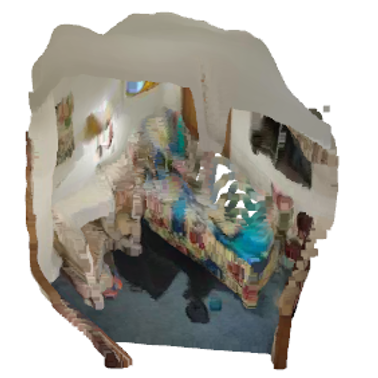}
\end{subfigure}
\end{minipage}
\vspace{3pt} \\

(d) &
\begin{minipage}[c]{0.19\textwidth}
\begin{subfigure}{1.0\textwidth}
  \centering
  \includegraphics[width=1\textwidth]{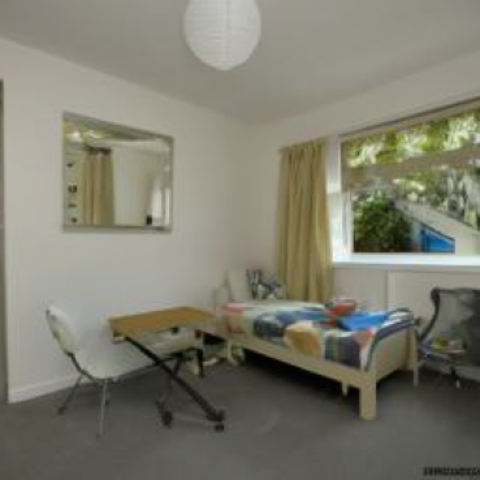}
\end{subfigure}
\end{minipage}
&
\begin{minipage}[c]{0.76\textwidth}
\begin{subfigure}{0.245\textwidth}
  \centering
  \includegraphics[width=1\linewidth]{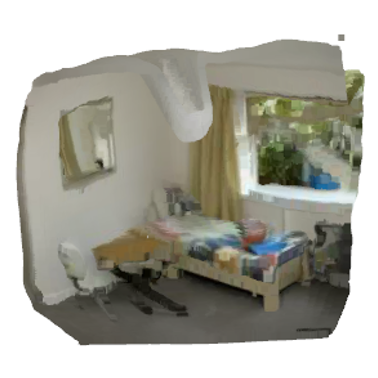}
\end{subfigure}
\begin{subfigure}{0.245\textwidth}
  \centering
  \includegraphics[width=1\linewidth]{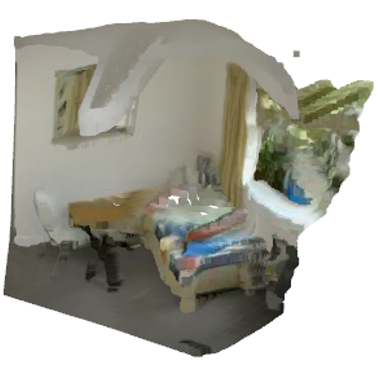}
\end{subfigure}
\begin{subfigure}{0.245\textwidth}
  \centering
  \includegraphics[width=1\linewidth]{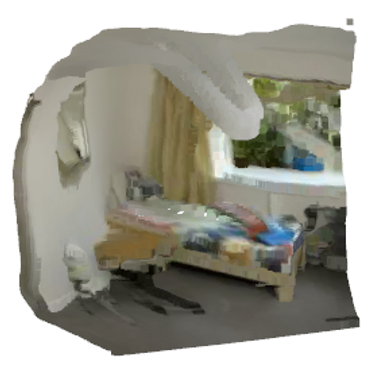}
\end{subfigure}
\begin{subfigure}{0.245\textwidth}
  \centering
  \includegraphics[width=1\linewidth]{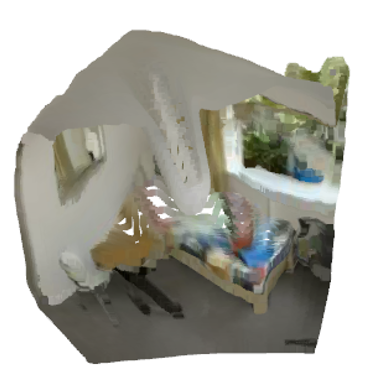}
\end{subfigure}
\end{minipage}
\vspace{3pt} \\

(e) &
\begin{minipage}[c]{0.19\textwidth}
\begin{subfigure}{1.0\textwidth}
  \centering
  \includegraphics[width=1\textwidth]{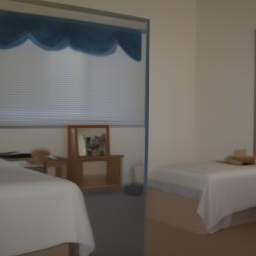}
\end{subfigure}
\end{minipage}
&
\begin{minipage}[c]{0.76\textwidth}
\begin{subfigure}{0.245\textwidth}
  \centering
  \includegraphics[width=1\linewidth]{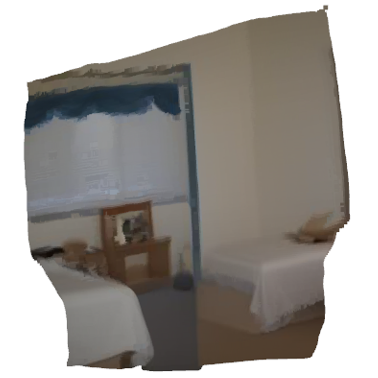}
\end{subfigure}
\begin{subfigure}{0.245\textwidth}
  \centering
  \includegraphics[width=1\linewidth]{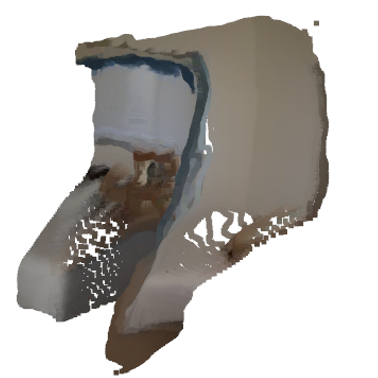}
\end{subfigure}
\begin{subfigure}{0.245\textwidth}
  \centering
  \includegraphics[width=1\linewidth]{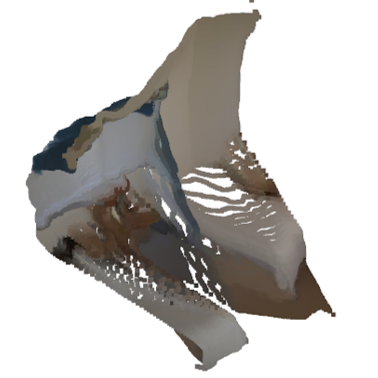}
\end{subfigure}
\begin{subfigure}{0.245\textwidth}
  \centering
  \includegraphics[width=1\linewidth]{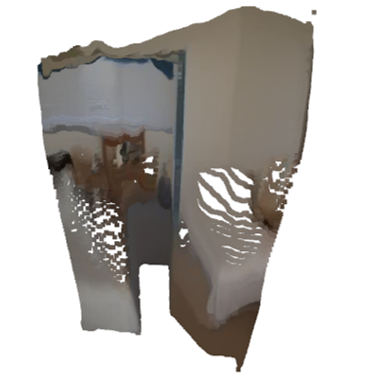}
\end{subfigure}
\end{minipage}
\vspace{3pt} \\

(f) &
\begin{minipage}[c]{0.19\textwidth}
\begin{subfigure}{1.0\textwidth}
  \centering
  \includegraphics[width=1\textwidth]{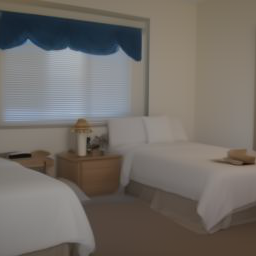}
\caption{synthesized image}
\end{subfigure}
\end{minipage}
&
\begin{minipage}[c]{0.76\textwidth}
\begin{subfigure}{0.245\textwidth}
  \centering
  \includegraphics[width=1\linewidth]{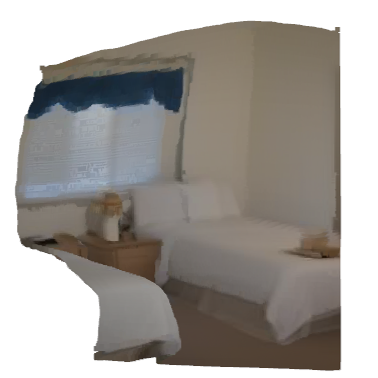}
\caption{front}
\end{subfigure}
\begin{subfigure}{0.245\textwidth}
  \centering
  \includegraphics[width=1\linewidth]{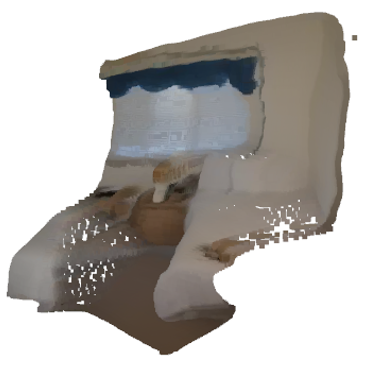}
\caption{right}
\end{subfigure}
\begin{subfigure}{0.245\textwidth}
  \centering
  \includegraphics[width=1\linewidth]{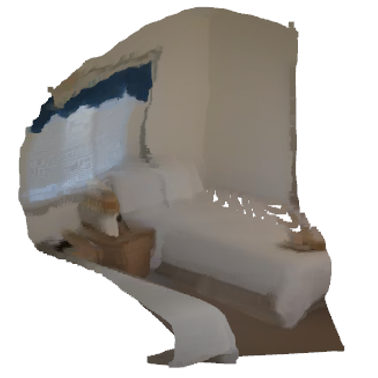}
\caption{left}
\end{subfigure}
\begin{subfigure}{0.245\textwidth}
  \centering
  \includegraphics[width=1\linewidth]{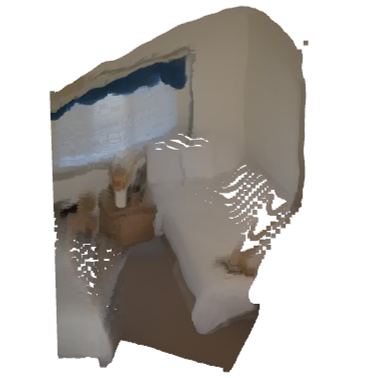}
\caption{up}
\end{subfigure}
\end{minipage} 
\end{tabular}
\vspace{-5pt}
\caption{\textbf{Visualization of point cloud representation obtained by depth information.} (a), (c), (e) are generated samples from ADM, and (b), (d), (f) are samples from ADM with our guidance method. The left column is generated image, and the other columns are point cloud visualizations from four different views.}
\label{fig:pcd}
\end{figure*}
\newpage

%% file: supple/4.Surface_normal.tex
\begin{figure}[p]
\centering
\hsize=\textwidth
\captionsetup[subfigure]{labelformat=empty}
\begin{subfigure}{.497\textwidth}
\centering
\lineskip=0pt
      \includegraphics[width=0.32\linewidth]{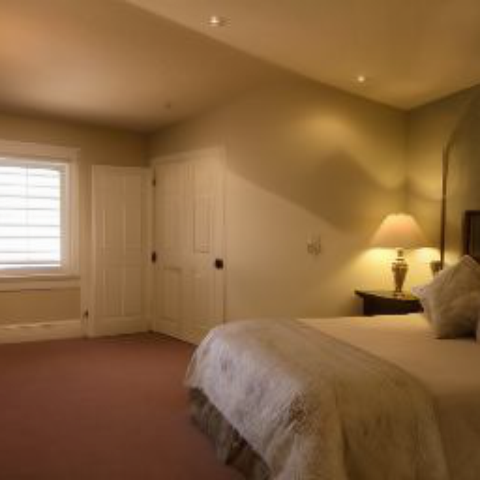}\hspace{-0.25em}
      \includegraphics[width=0.32\linewidth]{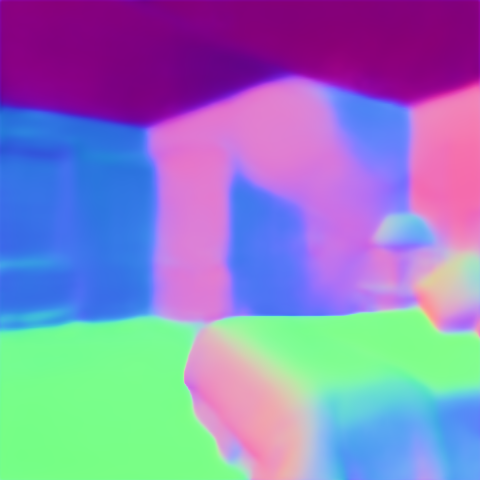}\hspace{-0.25em}
      \includegraphics[width=0.32\linewidth]{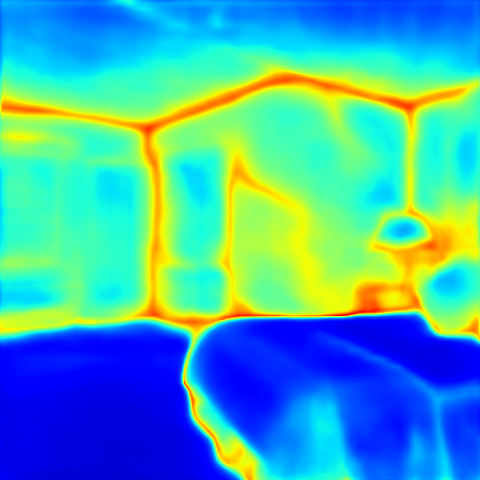}\\
      \includegraphics[width=0.32\linewidth]{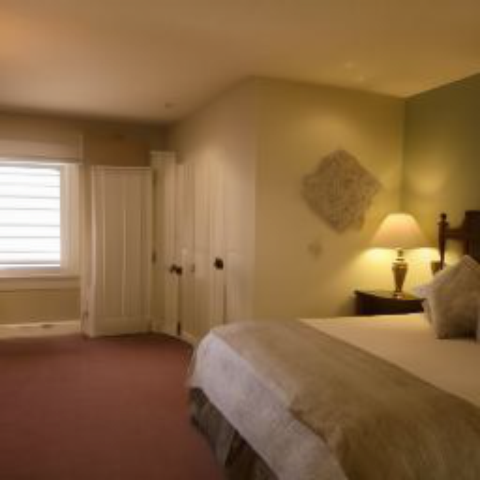}\hspace{-0.25em}
      \includegraphics[width=0.32\linewidth]{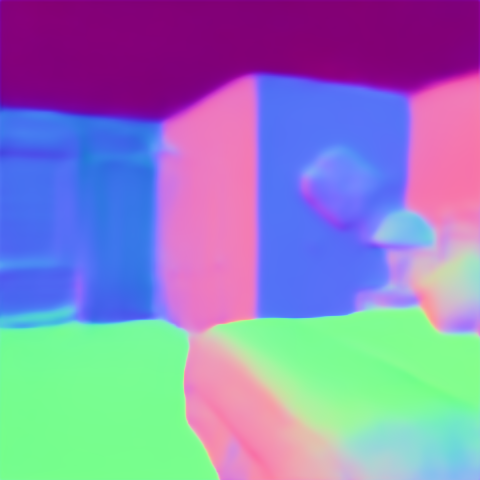}\hspace{-0.25em}
      \includegraphics[width=0.32\linewidth]{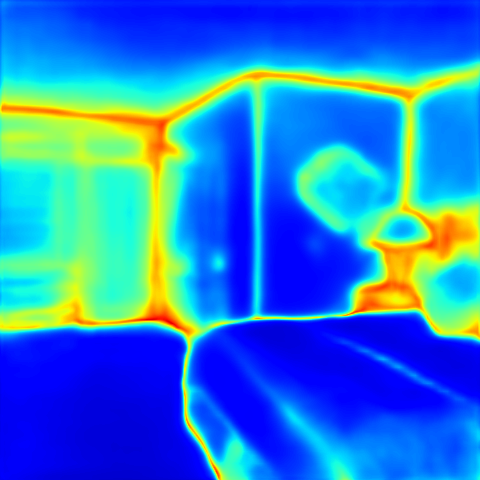}
\end{subfigure}\hspace{-1em}
\begin{subfigure}{.497\textwidth}
\centering
\lineskip=0pt
      \includegraphics[width=0.32\linewidth]{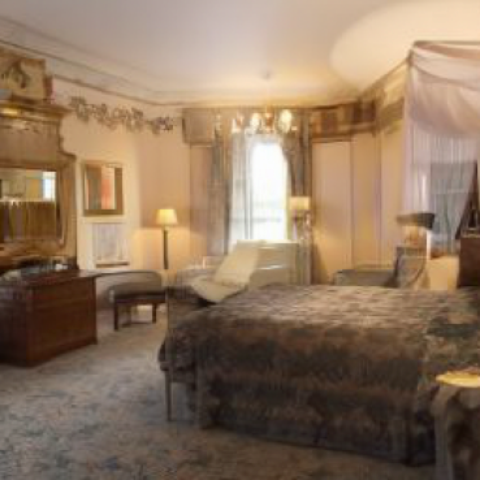}\hspace{-0.25em}
      \includegraphics[width=0.32\linewidth]{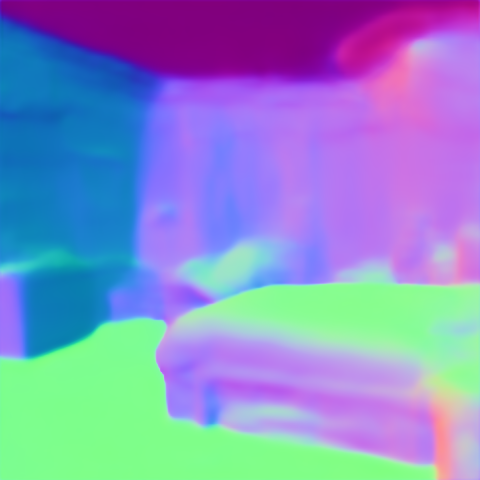}\hspace{-0.25em}
      \includegraphics[width=0.32\linewidth]{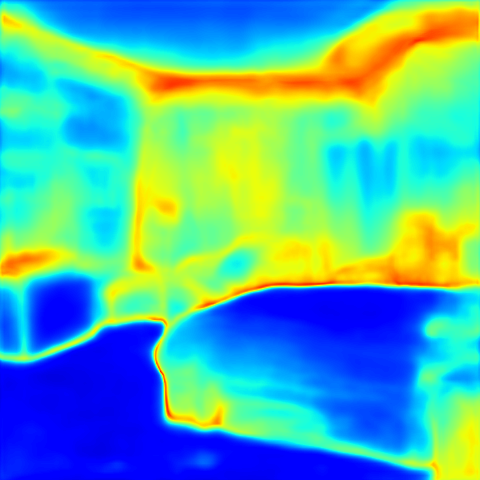}\\
      \includegraphics[width=0.32\linewidth]{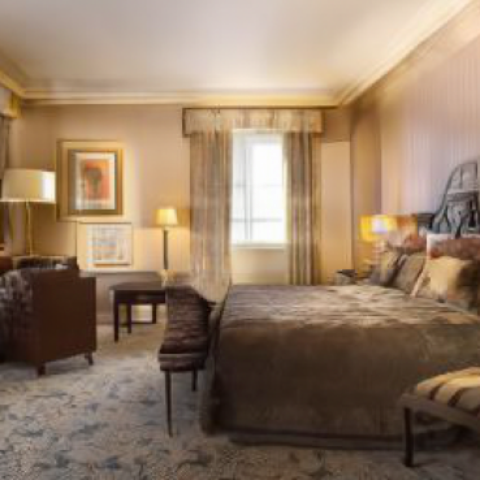}\hspace{-0.25em}
      \includegraphics[width=0.32\linewidth]{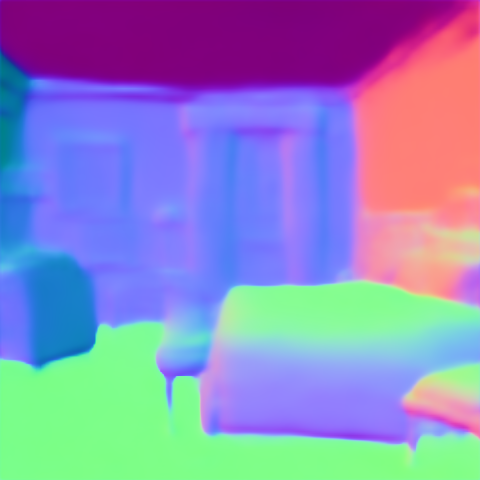}\hspace{-0.25em}
      \includegraphics[width=0.32\linewidth]{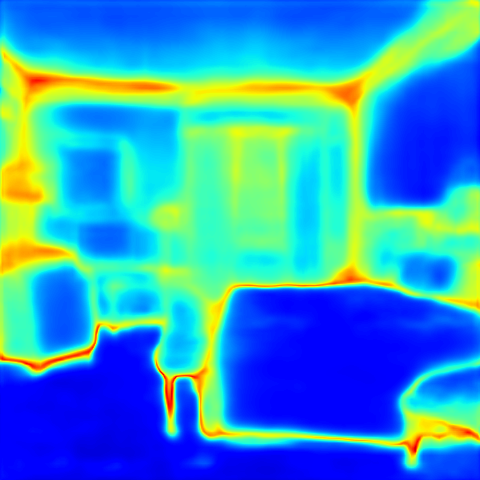}

\end{subfigure}
\\
\vspace{0.2em}
\begin{subfigure}{.497\textwidth}
\centering
\lineskip=0pt
      \includegraphics[width=0.32\linewidth]{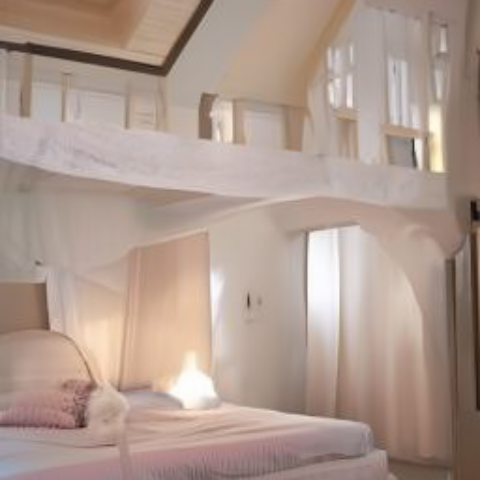}\hspace{-0.25em}
      \includegraphics[width=0.32\linewidth]{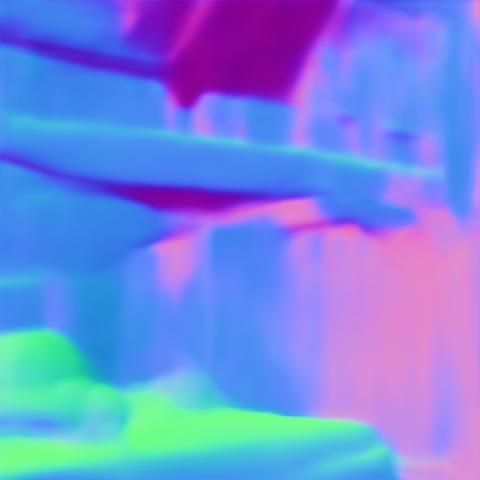}\hspace{-0.25em}
      \includegraphics[width=0.32\linewidth]{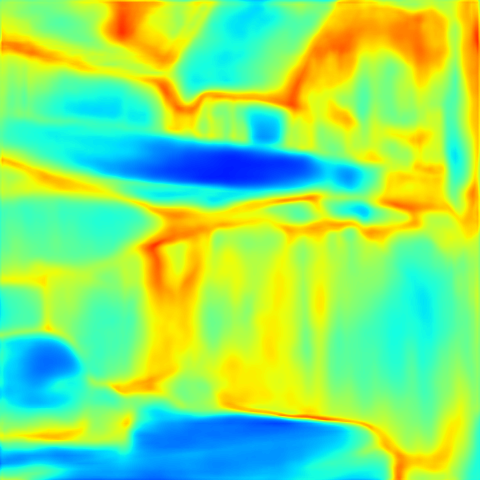}\\
      \includegraphics[width=0.32\linewidth]{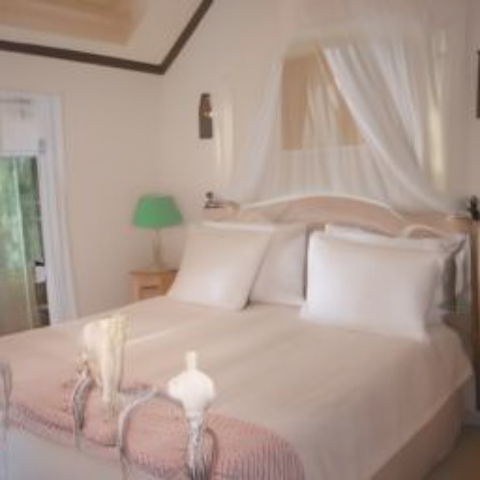}\hspace{-0.25em}
      \includegraphics[width=0.32\linewidth]{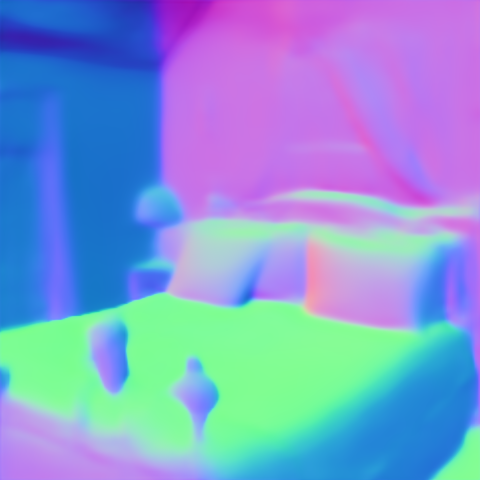}\hspace{-0.25em}
      \includegraphics[width=0.32\linewidth]{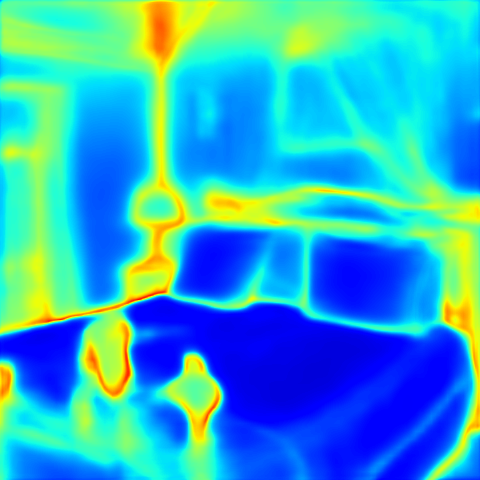}
\end{subfigure}\hspace{-1em}
\begin{subfigure}{.497\textwidth}
\centering
\lineskip=0pt
      \includegraphics[width=0.32\linewidth]{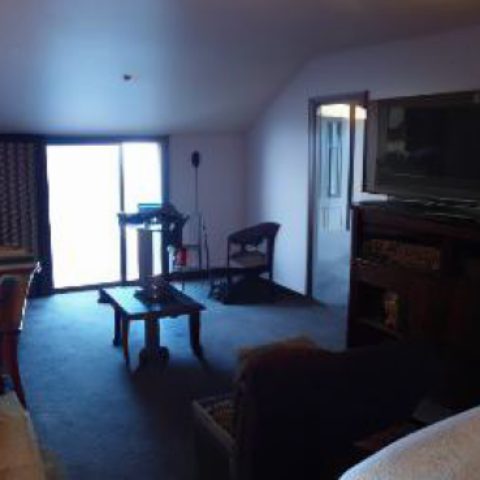}\hspace{-0.25em}
      \includegraphics[width=0.32\linewidth]{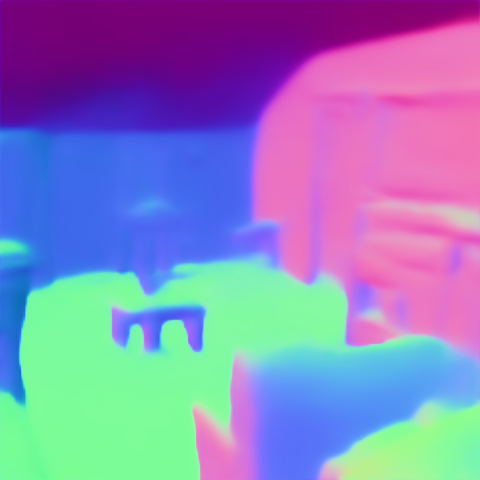}\hspace{-0.25em}
      \includegraphics[width=0.32\linewidth]{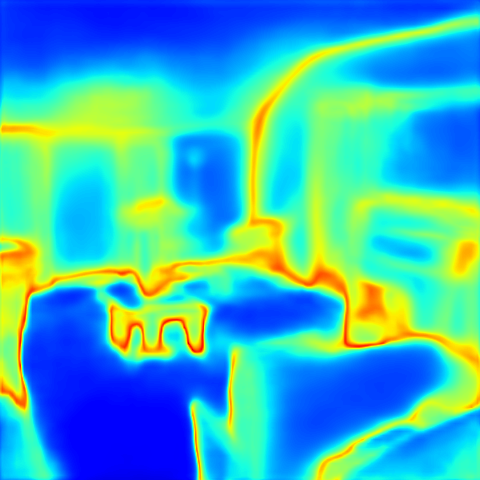}\\
      \includegraphics[width=0.32\linewidth]{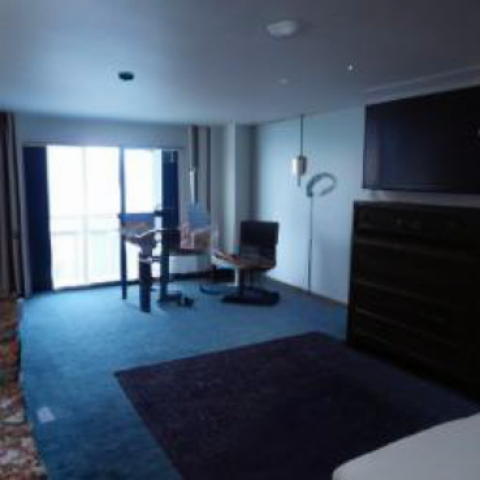}\hspace{-0.25em}
      \includegraphics[width=0.32\linewidth]{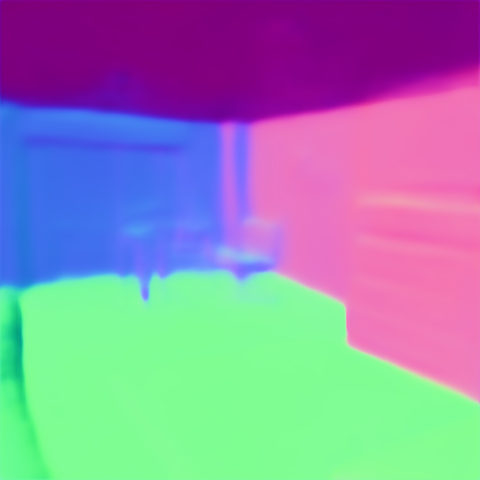}\hspace{-0.25em}
      \includegraphics[width=0.32\linewidth]{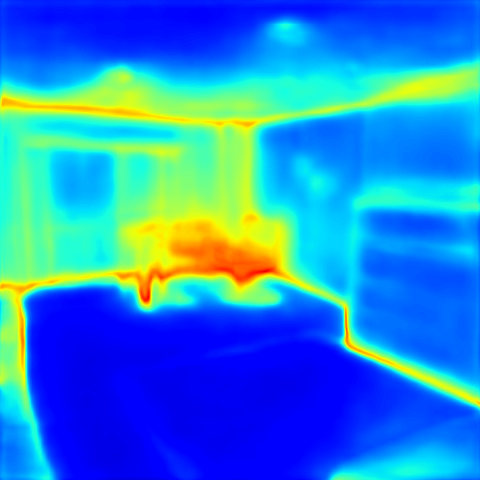}
\end{subfigure}
\\
\vspace{0.2em}
\begin{subfigure}{.497\textwidth}
\centering
\lineskip=0pt
      \includegraphics[width=0.32\linewidth]{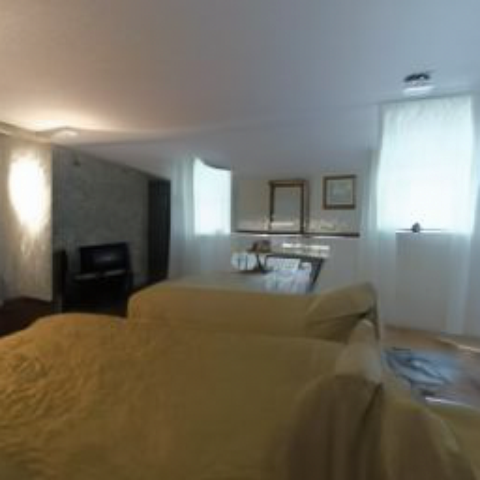}\hspace{-0.25em}
      \includegraphics[width=0.32\linewidth]{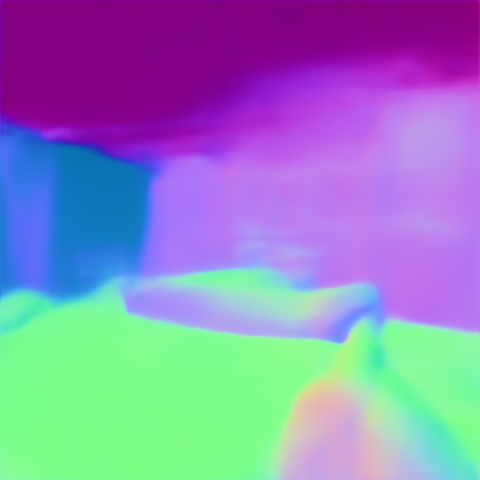}\hspace{-0.25em}
      \includegraphics[width=0.32\linewidth]{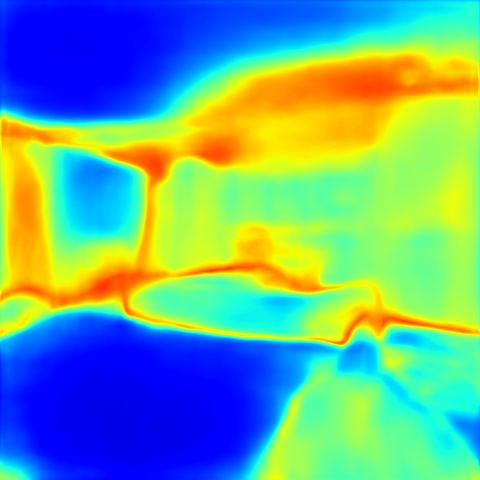}\\
      \includegraphics[width=0.32\linewidth]{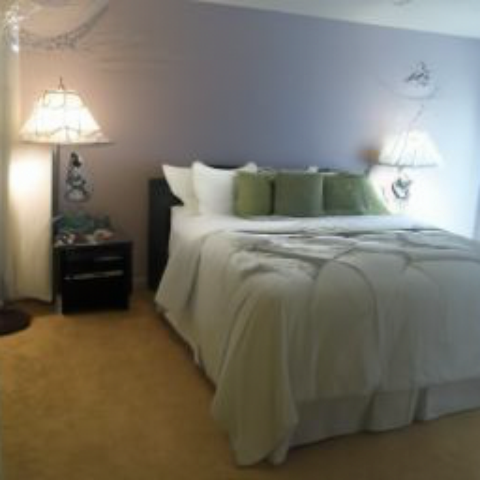}\hspace{-0.25em}
      \includegraphics[width=0.32\linewidth]{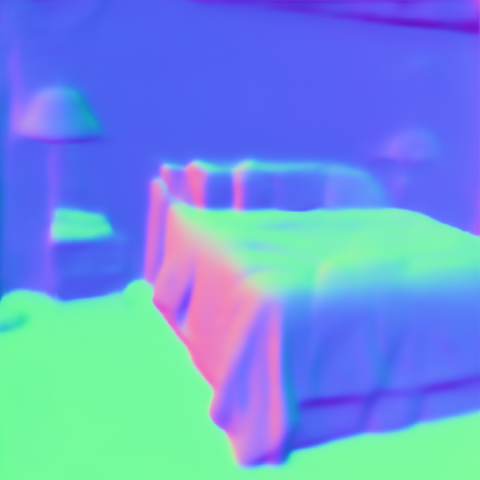}\hspace{-0.25em}
      \includegraphics[width=0.32\linewidth]{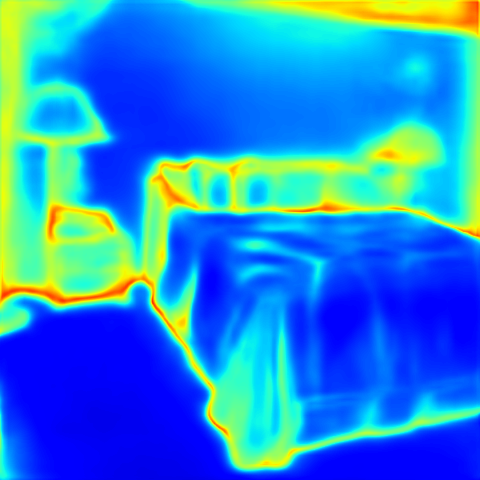}
\end{subfigure}\hspace{-1em}
\begin{subfigure}{.497\textwidth}
\centering
\lineskip=0pt
      \includegraphics[width=0.32\linewidth]{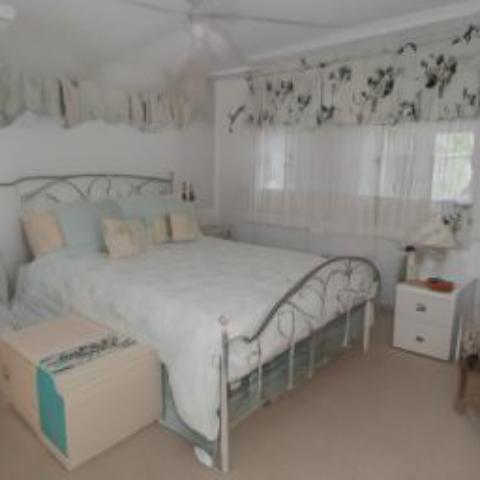}\hspace{-0.25em}
      \includegraphics[width=0.32\linewidth]{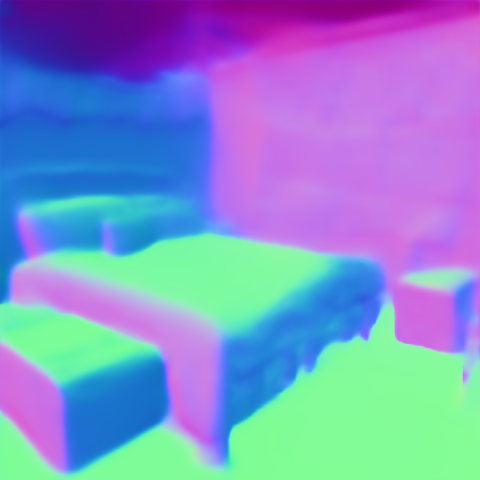}\hspace{-0.25em}
      \includegraphics[width=0.32\linewidth]{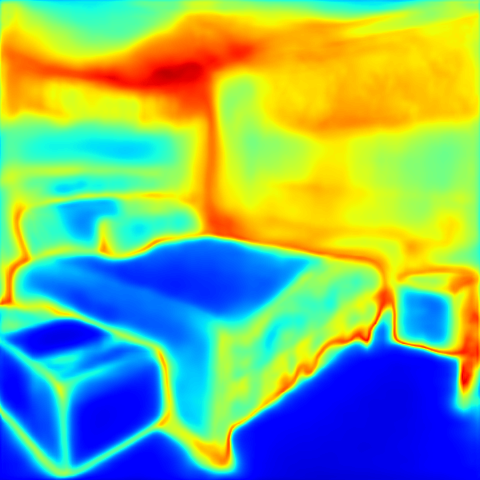}\\
      \includegraphics[width=0.32\linewidth]{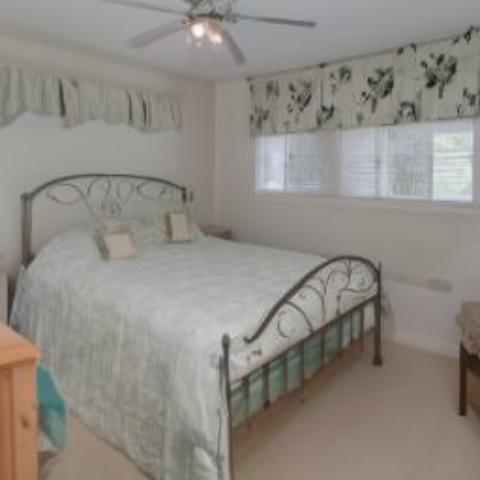}\hspace{-0.25em}
      \includegraphics[width=0.32\linewidth]{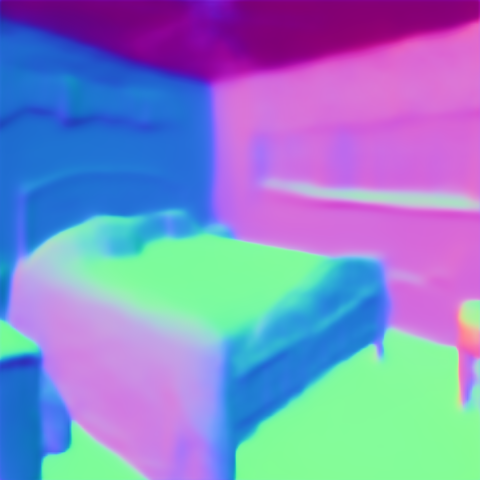}\hspace{-0.25em}
      \includegraphics[width=0.32\linewidth]{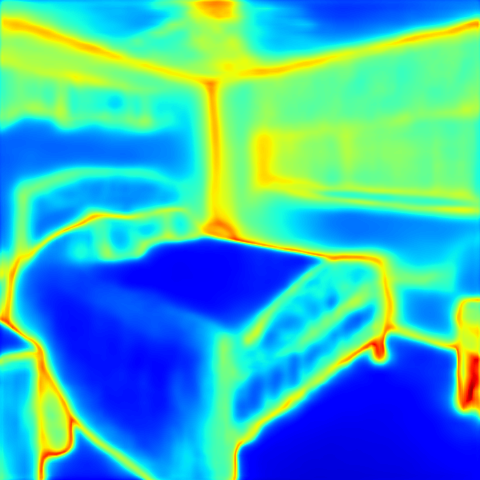}
\end{subfigure}
\\
\vspace{0.2em}

\begin{subfigure}{.497\textwidth}
\centering
\lineskip=0pt
      \includegraphics[width=0.32\linewidth]{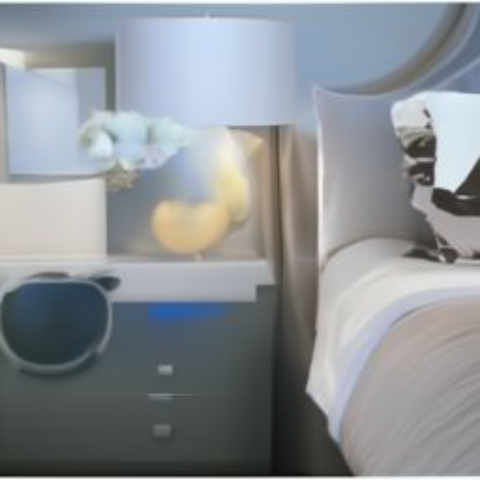}\hspace{-0.25em}
      \includegraphics[width=0.32\linewidth]{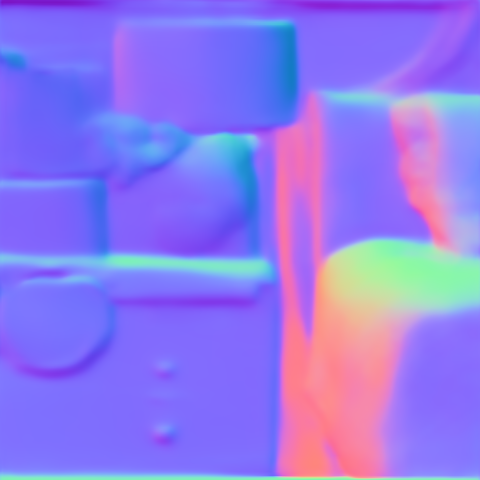}\hspace{-0.25em}
      \includegraphics[width=0.32\linewidth]{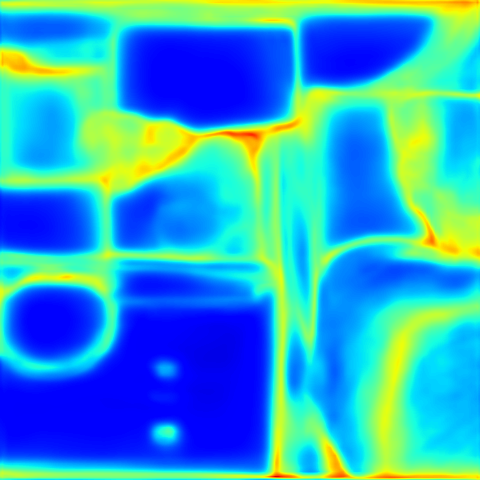}\\
      \includegraphics[width=0.32\linewidth]{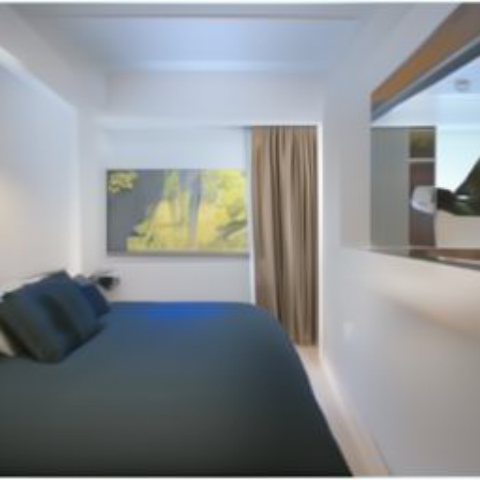}\hspace{-0.25em}
      \includegraphics[width=0.32\linewidth]{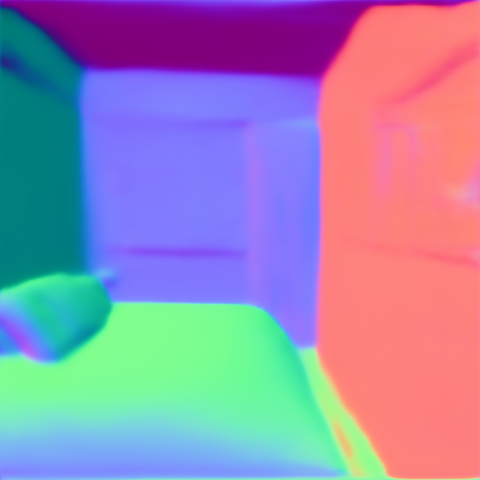}\hspace{-0.25em}
      \includegraphics[width=0.32\linewidth]{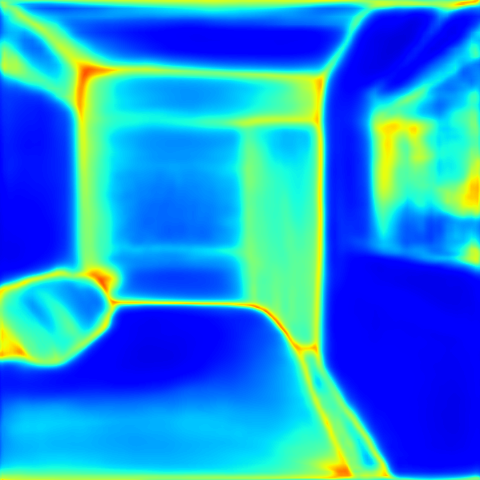}
\end{subfigure}\hspace{-1em}
\begin{subfigure}{.497\textwidth}
\centering
\lineskip=0pt
      \includegraphics[width=0.32\linewidth]{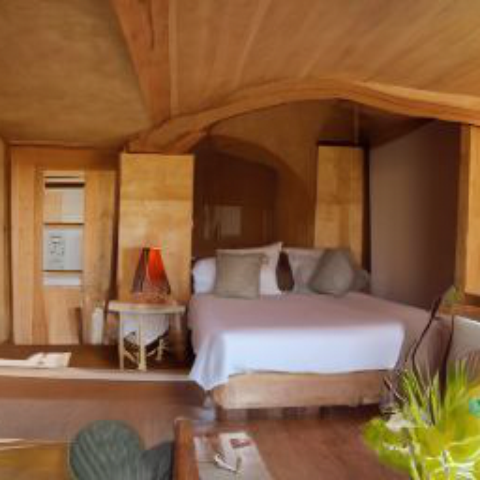}\hspace{-0.25em}
      \includegraphics[width=0.32\linewidth]{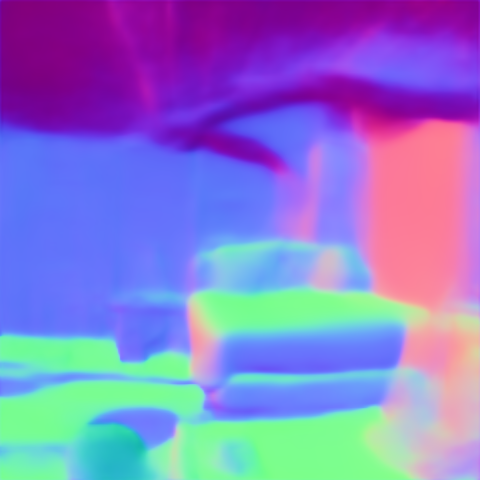}\hspace{-0.25em}
      \includegraphics[width=0.32\linewidth]{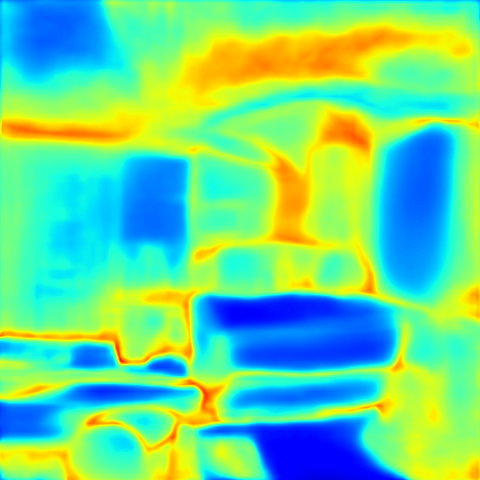}\\
      \includegraphics[width=0.32\linewidth]{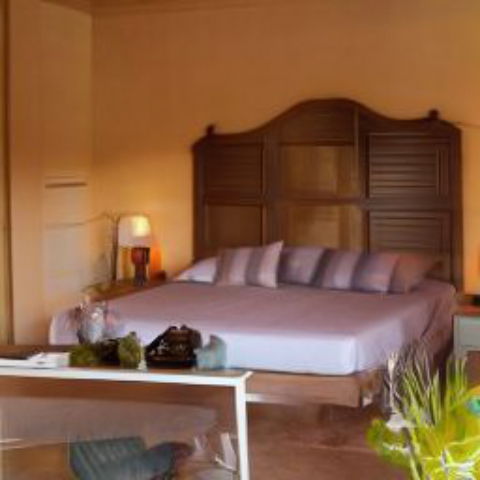}\hspace{-0.25em}
      \includegraphics[width=0.32\linewidth]{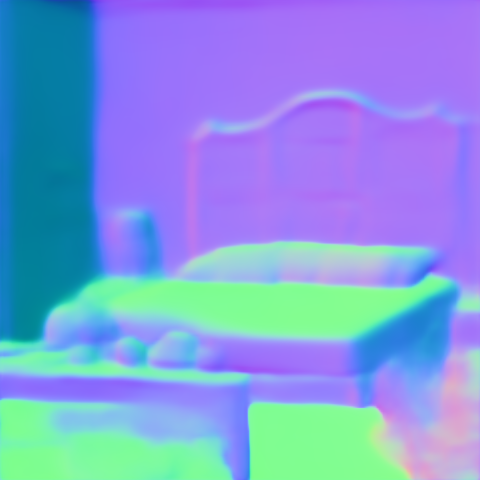}\hspace{-0.25em}
      \includegraphics[width=0.32\linewidth]{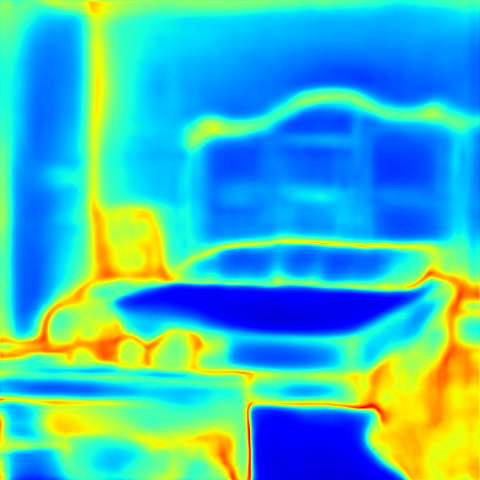}
\end{subfigure}

\vspace{-5pt}
\caption{\textbf{Qualitative comparison of the baseline and our method using surface normals and corresponding uncertainty.}}
\label{fig:normal}
\end{figure}
\newpage